\documentclass{article}

\usepackage{arxiv}
\usepackage[numbers]{natbib}

\usepackage[utf8]{inputenc} 
\usepackage[T1]{fontenc}    
\usepackage{hyperref}       
\usepackage{url}            
\usepackage{booktabs}       
\usepackage{nicefrac}       
\usepackage{microtype}      
\usepackage{lipsum}		
\usepackage{graphicx}
\usepackage{natbib}
\usepackage{doi}

\usepackage{amsmath,amssymb,amsfonts}
\usepackage{algorithmic}
\usepackage{graphicx}
\usepackage{textcomp}
\usepackage{amsthm}
\usepackage{url}

\usepackage[group-separator={,},group-minimum-digits=4]{siunitx}

\newtheorem{theorem}{Theorem}

\usepackage{amsmath,amssymb,amsfonts}
\usepackage{algorithmic}
\usepackage{graphicx}
\usepackage{textcomp}

\usepackage[utf8]{inputenc} 
\usepackage[T1]{fontenc}    
\usepackage{hyperref}       
\usepackage{url}            
\usepackage{booktabs}       
\usepackage{nicefrac}       
\usepackage{microtype}      
\usepackage{xcolor}         

\usepackage{multirow}

\usepackage{threeparttable}
\usepackage{lipsum}		
\usepackage{doi}

\usepackage{rotating}

\usepackage[table]{xcolor}
\usepackage{ulem}
\definecolor{deflightgray}{gray}{0.97}
\definecolor{defgray}{gray}{0.90}

\newcommand{\minitab}[2][l]{\begin{tabular}{#1}#2\end{tabular}}

\usepackage[utf8]{inputenc} 
\usepackage[T1]{fontenc}    
\usepackage{hyperref}       
\usepackage{url}            
\usepackage{booktabs}       
\usepackage{nicefrac}       
\usepackage{microtype}      
\usepackage{lipsum}		
\usepackage{graphicx}
\usepackage{natbib}
\usepackage{doi}

\usepackage{amsmath,amssymb,amsfonts}
\usepackage{algorithmic}
\usepackage{graphicx}
\usepackage{textcomp}
\usepackage{amsthm}
\usepackage{url}

\usepackage[group-separator={,},group-minimum-digits=4]{siunitx}

\usepackage{amsmath,amssymb,amsfonts}
\usepackage{algorithmic}
\usepackage{graphicx}
\usepackage{textcomp}

\usepackage[utf8]{inputenc} 
\usepackage[T1]{fontenc}    
\usepackage{hyperref}       
\usepackage{url}            
\usepackage{booktabs}       
\usepackage{nicefrac}       
\usepackage{microtype}      
\usepackage{xcolor}         

\usepackage{multirow}

\usepackage{threeparttable}
\usepackage{lipsum}		
\usepackage{doi}

\title{Efficient 3D affinely equivariant CNNs with weighted augmented spherical Fourier Bessel bases}

\author{Wenzhao Zhao	\thanks{Wenzhao Zhao is with School of Computer and Artificial Intelligence, Nanjing University of Finance and Economics and Interdisciplinary Center for Scientific Computing, Mannheim Institute for Intelligent Systems in Medicine, Medical Faculty Mannheim, Heidelberg University. 
		Steffen Albert and Frank G. Zöllner are with Computer Assisted Clinical Medicine, Mannheim Institute for Intelligent Systems in Medicine, Medical Faculty Mannheim, Heidelberg University.
		Barbara D. Wichtmann, Angelika Maurer and Ulrike Attenberger are with Department of Diagnostic and Interventional Radiology, University Hospital Bonn.
		Jürgen Hesser is with Interdisciplinary Center for Scientific Computing,
		Central Institute for Computer Engineering,
		CSZ Heidelberg Center for Model-Based AI, Data Analysis and Modeling in Medicine, Mannheim Institute for Intelligent Systems in Medicine, Medical Faculty Mannheim, Heidelberg University.  E-mail: zhaowenzhaoyz@163.com.
	}
	\And
	Steffen Albert 
	\And
	Barbara D. Wichtmann 
	\And	
	Angelika Maurer 
	\And
	Ulrike Attenberger 
	\And
	Frank G. Zöllner 
	\And  	
	Jürgen Hesser	
}

\renewcommand{\shorttitle}{3D affine equivariant CNN}

\begin{document}
\maketitle
\begin{abstract}
Filter-decomposition-based group equivariant convolutional neural networks (CNNs) have shown promising stability and data efficiency for 3D image feature extraction. However, these networks, which rely on parameter sharing and discrete transformation groups, often underperform in modern deep neural network architectures for processing volumetric images with dense 3D textures, such as the common 3D medical images.
To address these limitations, this paper presents an efficient non-parameter-sharing continuous 3D affine group equivariant neural network for volumetric images. This network uses an adaptive aggregation of Monte Carlo augmented spherical Fourier-Bessel filter bases to improve the efficiency and flexibility of 3D group equivariant CNNs for volumetric data.	
Unlike existing methods that focus only on angular orthogonality in filter bases, the introduced spherical Bessel Fourier filter base incorporates both angular and radial orthogonality to improve feature extraction.	
Experiments on four medical image segmentation datasets and two seismic datasets show that the proposed methods achieve better affine group equivariance and superior segmentation accuracy than existing 3D group equivariant convolutional neural network layers, significantly improving the training stability and data efficiency of conventional CNN layers (at 0.05 significance level). The code is available at \url{https://github.com/ZhaoWenzhao/WMCSFB}.
\end{abstract}


\section{Introduction}
\label{sec:introduction}
Convolution-based neural networks or CNNs have gained great popularity for image processing\cite{qiu2018dcfnet,jiao2019survey}. 
Convolutional layers are translation-equivariant because of their sliding window strategy\cite{fukushima1980neocognitron,lecun1989backpropagation}, which allows CNNs to learn translaton-equivariant features efficiently and stably. 
In addition to translation equivariance, the equivariance or invariance under other geometric transformations (such as rotation, scaling, or more generally, affine transformation) are also often desirable properties for many practical computer vision tasks\cite{cheng2019rotdcf,zhong2022pesa,mo2024ric,you2024mine,you2023rethinking}.
Extending the translation equivariance to equivariance over other transformation groups leads to the group equivariant CNNs (G-CNNs).
The G-CNNs are first proposed by Cohen and Welling\cite{cohen2016group}, where a 2D rotation equivariant CNN is designed based on the transformations of the same learnable convolutional kernels. A filter-decomposition-based approach called steerable CNN was developed later for roto-reflection group equivariance\cite{cohen2016steerable}. After that, the considered group equivariance is extended gradually to 2D scale equivariance\cite{sangalli2021scale,sosnovik2019scale,sosnovik2021disco,zhu2022scaling}, scale-rotation equivariance\cite{gao2021deformation}. 
However, the above-mentioned research works heavily rely on parameter-sharing to build expensive convolutional units. This increases the difficulty of building deeper neural networks. Also, such parameter-sharing group equivariance greatly limits the potential for developing more complicated group equivariant CNN, such as affine group equivariant CNN.
Recently, a 2D non-parameter-sharing affine group equivariant CNN has been proposed\cite{zhao2023adaptive}, which shows promising data and computation efficiency and high flexibility because its non-parameter-sharing group equivariance allows efficient and flexible integration with deep CNN architectures for practical applications.

This paper focuses on 3D filter decomposition-based group equivariant convolutional layers for volumetric image processing. It should be noted that there has been a lot of research on designing rotation equivariant networks specifically for 3D point cloud data\cite{thomas2018tensor,zhu2023e2pn,zhdanov2023implicit,shen2024rotation}, which may not be suitable for volumetric medical images.  
The first 3D group equivariant CNN for volumetric data is proposed in \cite{weiler20183d} for discrete 3D rotation group equivariance. This work is summarised in \cite{cesa2021program} to have a generally applicable method for rotation equivariant CNNs from arbitrary CNNs. 
Then, a partial differential equation based G-CNN\cite{he2023neural} is proposed for favourable spatially adaptive equivariant feature learning from 3D volumetric images.
Instead of developing group equivariance methods for a single CNN layer, there is also work on combining multiple G-CNN layers to achieve good group equivariance\cite{shen2024efficient}. However, this is not the focus of this paper. 
The existing 3D G-CNN layers rely on parameter sharing and build computationally expensive convolutional layers for good performance, which limits their application to deep CNN architectures. In addition, they are limited to discrete rotation equivariant CNNs and mostly use simple spherical harmonic bases as filter bases, which only parameterise filters supported on a compact sphere to represent angular coordinates, but lack radial orthogonality.

In this paper, we propose an efficient non-parameter-sharing 3D affine group equivariant convolutional layer. Specifically, our
contributions are embodied in three aspects:
\begin{itemize}
	\item We propose a $GL^+(3,\mathbb{R})$ continuous affine group (a Lie group) equivariant CNN layer based on a decomposition of the 3D transformation matrix, which is efficiently implemented using Monte Carlo weighted group equivariant neural networks.
	
	\item Instead of the usual simple spherical harmonics, we introduce a more expressive basis by combining angular orthogonality and radial orthogonality, resulting in the spherical Fourier-Bessel bases, to improve the performance of 3D group equivariant neural networks.
	
	\item We demonstrate the use of the proposed methods to improve the training stability and data efficiency of state-of-the-art 3D deep CNNs in 3D medical image segmentation tasks combined with data augmentation.

\end{itemize}

In the following parts of this paper, we detail our methods in the Methods section and show the experiments and discussions in the Experiments section. The paper is summarised in the Conclusion section.

\section{Background}

This paper focuses on the development of equivariant convolutional layers. And to test the group equivariance, we choose the dense prediction task, volumetric image segmentation, as a test task (while image classification is suitable for testing group invariance). The tested or compared convolutional layers are used as the basic "building blocks" of the deep CNN models.

A deep CNN model embodies a way of stacking convolutional layers, i.e. the architecture of the deep CNN (ADCNN).
Among the various ADCNNs, Unet\cite{ronneberger2015u} is the most successful for both 2D and 3D image processing\cite{isensee2021nnu}. 
Along with the rise of transformer neural network architectures and other architectures\cite{U-Mamba} in recent years, CNNs remain the state-of-the-art deep learning neural networks for image processing by modernising its ADCNN accordingly\cite{hatamizadeh2022unetr}. ConvNeXt\cite{liu2022convnet} is the classic ADCNN that builds a modernised bottleneck architecture inspired by the architectures of transformers for 2D image feature extraction. More recently, inspired by ConvNeXt, a 3D version of ConvNeXt, called MedNeXt\cite{roy2023mednext}, has been developed to achieve state-of-the-art performance for 3D medical image segmentation. In this paper, we adopt this ADCNN as base models for convenience and our focus is on developing favourable basic affine equivariant convolutional layers that are used for building these deep CNNs. 

Apart from ADCNN, it is also found that the strategies for organizing and training different deep neural networks also strongly affect the performance \cite{chen2024hidiff,you2024mine,you2023rethinking,you2023bootstrapping,you2023implicit,you2023action++,10.5555/3600270.3602415}. For example, denoising diffusion strategies\cite{chen2024hidiff} have shown great advantage for excellent performance in image processing, but they significantly increase the inference time of neural networks. In this work, we do not consider denoising diffusion strategies for a convenient comparison of different group equivariant CNN layers. Instead, we consider the classical training strategy of nnUnet\cite{isensee2021nnu} as the basic setting for a fair and convenient comparison and testing of methods.

\section{Methods}
In this section, we will introduce the 3D weighted Monte Carlo group convolutional networks and spherical Fourier-Bessel basis in detail. To better understand our approach and related concepts, mathematical knowledge in abstract algebraic structures and representation theory will be helpful, which can be found in \cite{dummit2004abstract,gurarie2007symmetries}.

\subsection{Group and group equivariance}

In this paper, we consider the 3D positive general linear group $GL^+(3,\mathbb{R})$ and its subgroups. The 3D positive general linear group corresponds to 3D linear transformations that preserve the orientation, lines, and parallelism of 3D space. We consider the 3D transformation matrix $M(a)$ of any subgroup of $GL^+(3,\mathbb{R})$, where $a\in \mathbb{R}^d$ and $d$ is the number of parameters of the resulting transformation matrix. $M(a)$ can be any simple affine transformation ($d=1$):

\begin{equation}
	A_1(\alpha) = 
	\begin{bmatrix}
		2^{\alpha} & 0 & 0\\
		0 & 1 & 0\\
		0 & 0 & 1
	\end{bmatrix},
	\label{eq:scaling1}
\end{equation}	

\begin{equation}
	A_2(\beta) = 
	\begin{bmatrix}
		1 & 0 & 0\\
		0 & 2^{\beta} & 0\\
		0 & 0 & 1
	\end{bmatrix},
	\label{eq:scaling2}
\end{equation}	

\begin{equation}
	A_3(\gamma) = 
	\begin{bmatrix}
		1 & 0 & 0\\
		0 & 1 & 0\\
		0 & 0 & 2^{\gamma}
	\end{bmatrix},
	\label{eq:scaling3}
\end{equation}

\begin{equation}
	R_1(\theta_1) = 
	\begin{bmatrix}
		1 & 0 & 0 \\
		0 & \cos{\theta_1} & -\sin{\theta_1}\\
		0 & \sin{\theta_1} & \cos{\theta_1}
	\end{bmatrix},
	\label{eq:rotation1}
\end{equation}	

\begin{equation}
	R_2(\theta_2) = 
	\begin{bmatrix}        
		\cos{\theta_2} & 0 & -\sin{\theta_2}\\
		0 & 1 & 0 \\
		\sin{\theta_2} & 0 &\cos{\theta_2}
	\end{bmatrix},
	\label{eq:rotation2}
\end{equation}

\begin{equation}
	R_3(\theta_3) = 
	\begin{bmatrix}
		\cos{\theta_3} & -\sin{\theta_3}&0\\
		\sin{\theta_3} & \cos{\theta_3}&0\\
		0 & 0 & 1 
	\end{bmatrix},
	\label{eq:rotation3}
\end{equation}

\begin{equation}
	S_{01}(s_{01}) = 
	\begin{bmatrix}
		1 & s_{01} & 0\\
		0 & 1 & 0\\
		0 & 0 & 1
	\end{bmatrix},
	\label{eq:shear01}
\end{equation}

\begin{equation}
	S_{02}(s_{02}) = 
	\begin{bmatrix}
		1 & 0 & s_{02}\\
		0 & 1 & 0\\
		0 & 0 & 1
	\end{bmatrix},
	\label{eq:shear02}
\end{equation}
\begin{equation}
	S_{12}(s_{12}) = 
	\begin{bmatrix}
		1 & 0 & 0\\
		0 & 1 & s_{12}\\
		0 & 0 & 1
	\end{bmatrix},
	\label{eq:shear12}
\end{equation}
\begin{equation}
	S_{10}(s_{10}) = 
	\begin{bmatrix}
		1 & 0 & 0\\
		s_{10} & 1 & 0\\
		0 & 0 & 1
	\end{bmatrix},
	\label{eq:shear10}
\end{equation}

\begin{equation}
	S_{20}(s_{20}) = 
	\begin{bmatrix}
		1 & 0 & 0\\
		0 & 1 & 0\\
		s_{20} & 0 & 1
	\end{bmatrix},
	\label{eq:shear20}
\end{equation}

\begin{equation}
	S_{21}(s_{21}) = 
	\begin{bmatrix}
		1 & 0 & 0\\
		0 & 1 & 0\\
		0 & s_{21} & 1
	\end{bmatrix}.
	\label{eq:shear21}
\end{equation}

We can also construct a $M(a)$ by matrix decomposition to include all the elements in $GL^+(3,\mathbb{R})$. The decomposition is done by a series of multiplications of simple transformation matrices, for example,
\begin{equation}
	\begin{aligned}
		M(a)=R_1(\theta_1)R_3(\theta_3)\cdot \\
		A_1(\alpha)A_2(\beta)A_3(\gamma)\cdot\\
		S_{20}(s_{20})S_{10}(s_{10})S_{21}(s_{21})S_{01}(s_{01})\cdot\\
		S_{12}(s_{12})S_{02}(s_{02})
	\end{aligned}
	\label{eq:matrixdecomp}
\end{equation}	
where we have $a = (\theta_1,\theta_3,\alpha,\beta,\gamma, s_{01},s_{10},s_{02},s_{20},s_{12},s_{21})\in \mathbb{R}^{11}$ a parameter vector for transformations.

We have the following theorem for the above construction:
\begin{theorem}
	Any element in $GL^+(3,\mathbb{R})$ can be decomposed into the form of \eqref{eq:matrixdecomp}. Any $3\times 3$ matrix constructed via \eqref{eq:matrixdecomp} belongs to $GL^+(3,\mathbb{R})$.
	\label{theorem:decomp}
\end{theorem}

The proof of this theorem can be found in Appendix A. This theorem shows that such a decomposition builds the group $GL^+(3,\mathbb{R})$. As shown in the proof, such a decomposition can be redundant, since a rotation can be decomposed into shear matrices and scaling matrices. In this paper we adopt this decomposition for a convenient implementation that includes control parameters for the common types of simple transformations (i.e., rotation, scaling, and shear transformations).

A 3D affine group $G$ is constructed with the space vector $\mathbb{R}^3$ and the general linear group. Specifically, an affine group element $g\in G$ can be written as $g=(x, M(a))$ with $x\in \mathbb{R}^3$ the spatial position.
For any group element $g_1$ and $g_2$, the group product is defined as 
\begin{equation}
	\begin{aligned}
		g_1 \cdot g_2 & = (x_1, M(a_1)) \cdot (x_2, M(a_2))\\
		& = (x_2+M(a_1)x_2+x_1,M(a_1)M(a_2))        
	\end{aligned}
	\label{eq:groupproduct}
\end{equation}
The group product means the composition of two affine transformations.
In this paper, the group we are considering is assumed to be locally compact.

The application of a group transformation to an index set $\mathcal{X}$ is called a group action $T:G\times \mathcal{X}\rightarrow \mathcal{X}$.  The group product with respect to the group action satisfies the compatibility
\begin{equation}
	T(g_1 \cdot g_2,x) = T(g_1, T(g_2,x))
	\label{eq:groupproduct_gact}
\end{equation}

As for applying a group transformation to a function space $L_{V}(\mathcal{X}):\{f: \mathcal{X}\rightarrow V\}$, the group action $\mathbb{T}_g$ is defined accordingly as $\mathbb{T}_g: f\rightarrow f^\prime$  where $f^\prime (T(g,x))=f(x)$.

Finally, based on the above concepts, we can have a formal definition of group equivariance. Specifically, a group equivariance mapping $\phi: L_{V_1}(\mathcal{X}_1)\rightarrow L_{V_2}(\mathcal{X}_2)$ satisfies
\begin{equation}
	\forall g\in G, \phi(\mathbb{T}_g(f))=\mathbb{T}^{\prime}_g(\phi(f)) 
	\label{eq:equivariant}
\end{equation}	 
where $\mathbb{T}_g$ and $\mathbb{T}_g^{\prime}$ represents the group action on function $f$ and $f^{\prime}$, respectively. In other words, a group equivariant mapping means that the transformation of its inputs has the same effect as the transformation of the outputs whose corresponding inputs do not undergo any transformation.

\subsection{Group convolution}
According to \cite{kondor2018generalization}, we have
\begin{theorem}
	A feed-forward neural network is equivariant to the action of a compact group G on its inputs if and only if each layer of the network implements a generalized form of
	convolution derived from the equation below
	\begin{equation}
		(\psi*f)(g)=\int_{G}\psi(g^{-1}\cdot g^\prime)f(g^\prime)d\mu (g^\prime)
		\label{eq:equivariant_conv}
	\end{equation}	
	where $*$ is the convolution symbol, $\mu$ is the Haar measure, and $f,\psi: G\rightarrow \mathbb{R}$.
	\label{theorem:groupconv}
\end{theorem}
The convolution defined by \eqref{eq:equivariant_conv} is thereby called group convolution, which shows favorable group equivariance in lots of recent research works\cite{gao2021deformation,sosnovik2019scale}. 
Essentially, the transformation group convolution introduces a transformation-level averaging procedure to enhance the transformation equivariance or invariance of the mapping. Therefore, in this work, we assume the general effectiveness of the group convolution operator for any kind of geometric group and its common subsets. 

\subsection{Weighted Monte Carlo group convolutional network}
This work adopts the weighted Monte Carlo G-CNN (WMCG-CNN) strategy as in \cite{zhao2023adaptive}, which can be considered as a "preconditioning" to allow the neural network to have a good group equivariance at the starting point of the training with randomly initialized trainable weights $w$. Specifically, given $g = (x,M(a))$ and $g^\prime = (u,M(b))$, weighted Monte Carlo group convolution considers weighted group convolution integration on $\mathbb{R}\times G$:
\begin{equation}
	\begin{array}{l}
		f^{(l+1)}(g) \\
		= \int_{\mathbb{R}\times G} w\cdot \psi(g^{-1} \cdot g^\prime)f^{(l)}(g^\prime)d\mu (g^\prime)d\mu_w(w) 
	\end{array}
	\label{eq:wmcg}
\end{equation}	
where $f^{(l)}$ denotes the feature map of the $l$-th layer, and $\psi$ denotes the convolutional filter basis. Unlike equation \eqref{eq:equivariant_conv}, an additional integration along the dimension $\mathbb{R}$ is added.

Let $g=(x,M(a))$ and $g^\prime=(u,M(b))$.
We rewrite equation \eqref{eq:wmcg} as
\begin{equation}
	\begin{array}{l}
		f_{dec}^{(l+1)}(x,a)  \\
		=  \int_{\mathbb{R}}\int_{\mathbb{R}^d} \int_{\mathbb{R}^3} w C(b)\psi(-x+M(-a)u,M(-a)M(b)) \\f^{(l)}(u,b)dudbdw
	\end{array}
	\label{eq:wmcg_d}
\end{equation}	
where $d$ the number of variables in the parameter vector $a$ or $b$.
$C(b)$ is the normalization coefficient with respect to the Haar measure $\mu (g^\prime)$. For the affine matrix $M(b)$ involving scaling transforms, there is $C(b)=2^{-2\alpha_b-2\beta_b-2\gamma_b}$.

Since we focus on the group equivariance of the convolutional layers, also for simplicity, here the constant coefficients, bias terms, and point-wise nonlinearity in neural networks are omitted.

\begin{figure*}[!t]
	\centerline{\includegraphics[width=14.0cm]{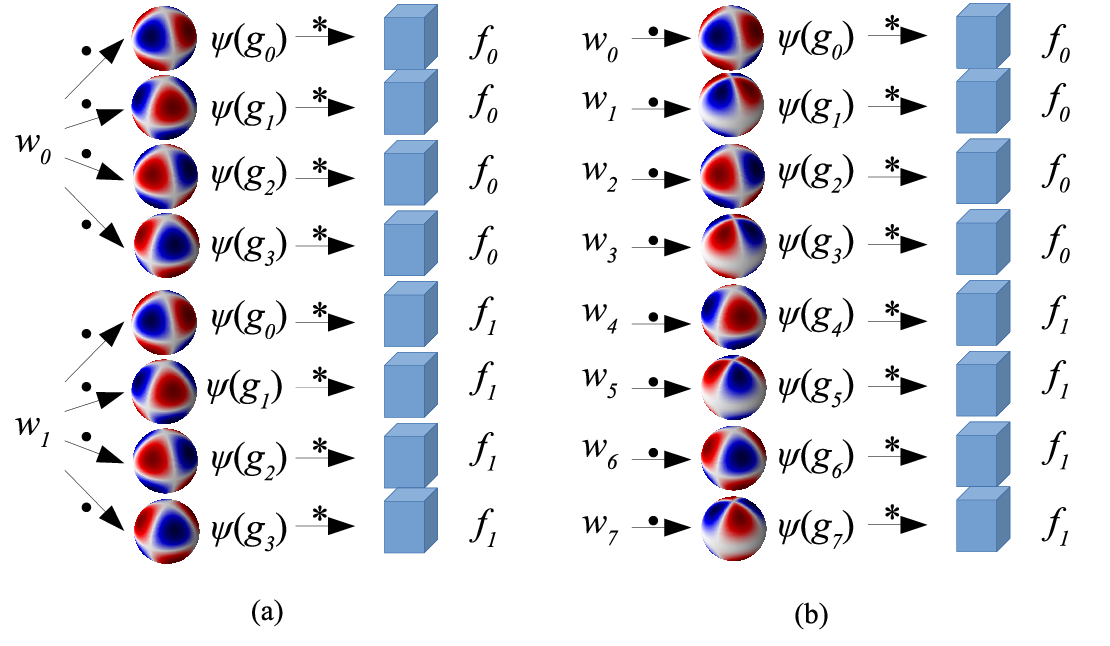}}
	\caption{The comparison of group equivariant convolutions, where $\cdot$ means multiplication and $*$ means convolution. $w_i$ denotes the $i$-th learnable weight, $\psi(g_i)$ denotes the filter augmented by the $i$-th transformation group member $g_i$, and $f_i$ denotes the $i$-th feature map. (a) The parameter-sharing strategy for common G-CNNs\cite{kondor2018generalization,weiler20183d,cesa2021program}; (b) The WMCG-CNN\cite{zhao2023adaptive}.}
	\label{fig_wgcnn}
\end{figure*}

The implementation of the discretization of the continuous transform is based on a Monte Carlo integral approximation. Unlike conventional G-CNN, WMCG-CNN eliminates the weight-sharing between feature channels to achieve an adaptive fusion of these Monte Carlo augmented filters(or feature maps). Specifically, let $c_o$ and $c_i$ be the channel numbers of output and input feature maps, respectively.
For each input-output channel pair, we draw randomly a sample of $b$ and a sample of weight $w$. For each output channel, we have a sample of $a$. For each sample of $a$, there are $N$ samples of $b$. Then we have the discretization implementation of equation \eqref{eq:wmcg_d} written as

\begin{equation}
	\begin{array}{l}
		f^{(l+1)}_{c_o}(x,a_{c_o}) =\sum_{c_i=0}^{N-1}\sum_{u} w^{(l)}_{c_o,c_i}C(b)\cdot 
		\\ \psi(-x+M(-a_{c_o})u,M(-a_{c_o})M(b_{b_o,c_i})) f^{(l)}_{c_i}(u,b_{c_o,c_i})		
	\end{array}
	\label{eq:gcnn}
\end{equation}	
A visual demonstration of this group convolution for WMCG-CNN is shown in Fig. \ref{fig_wgcnn}, with $N=4$, and $C(b)$ and $\sum_{c_i=0}^{N-1}$ omitted for simplicity.

Replacing the weighted function $\psi$ in \eqref{eq:gcnn} with the weighted sum of multiple filter bases $W^{(l)}_{c_o,c_i}(x,M(a)) = \sum_j w^{(l)}_{c_o,c_i,j} \Tilde{\psi}_j(x,M(a)) $ with $j$ the basis number and $\Tilde{\psi}$ the $j$-th basis, we get the fully fledged WMCG-CNN as
\begin{equation}
	\begin{array}{l}
		f^{(l+1)}_{c_o}(x,a_{c_o}) =\sum_{c_i=0}^{N-1} \sum_{u} C(b) W^{(l)}_{c_o,c_i}(\\ 
		-x+M(-a_{c_o})u,M(-a_{c_o})M(b_{c_o,c_i})) f^{(l)}_{c_i}(u,b_{c_o,c_i})
		,
	\end{array}
	\label{eq:mcgcnn2}
\end{equation}	
As suggested in \cite{zhao2023adaptive}, the WMCG-CNN is often followed by scalar convolutional layers ($1\times 1\times 1$ convolutional kernel for 3D CNN) to improve its equivariance performance.

In the inference phase, by pre-computing the convolutional kernel $W$ in \eqref{eq:mcgcnn2},  the WMCG-CNNs have the same computational burden as the standard CNN when using the same kernel size. During the training phase, there is usually a tiny increase in computational burden due to the weighted sum of the filter bases.

\subsection{Spherical Fourier-Bessel basis}
Considering a spherical coordinate, the existing works on SE(3) G-CNN\cite{weiler20183d,cesa2021program} for volumetric data use the following filter bases
\begin{equation}
	\psi_{lmn}(\mathbf{r})=\Phi(r-n)Y_{lm}(\theta,\phi)
	\label{eq:sph}
\end{equation}
where $\mathbf{r}=(r,\theta,\phi)$, the Gaussian function $\Phi(r-n)=\exp(-\frac{1}{2}(r-n)^2/\sigma^2)$, and $Y_{lm}(\theta,\phi)$ are the spherical harmonics.

It is clear that the spherical harmonics satisfy angular orthogonality, which means
\begin{equation}
	\int_{0}^{2\pi}\int_{0}^{\pi} Y_{l_1,m_1}(\theta,\phi)Y^{*}_{l_2,m_2}(\theta,\phi) sin(\theta) d\theta d\phi = \delta_{m_1,m_2}\delta_{l_1,l_2}
	\label{eq:sph_ortho}
\end{equation}
where $\delta$ is the Kronecker delta symbol.

However, the radial part $\Phi(r-n)$ lacks orthogonality. 
It is believed that the orthogonality of the bases allows a compact "good" representation of local features by using a small number of bases\cite{xie2018orthogonality}.
Following \cite{binney1991gaussian,fisher1995wiener}, we replace the Gaussian radial function with the spherical Bessel function to obtain the spherical Fourier-Bessel (sFB) basis

\begin{equation}
	\psi_{lmn}(\mathbf{r})=j_{l}(k_n r)Y_{lm}(\theta,\phi)
	\label{eq:sFB}
\end{equation}
where $j_{l}(\cdot)$ is the spherical Bessel function. The index $k_n$ denotes the discrete spectrum of the radial modes. As in \cite{fisher1995wiener}, the value of $k_n$ satisfies the continuous boundary condition by setting $j_{l-1}(k_n R) = 0$ for all $l$, where $R$ is the maximum radial of the kernel coordinate.

The Bessel function satisfies angular orthogonality 
\begin{equation}
	\int_{0}^{R} r^2 j_{l}(k_n r) j_{l}(k_{n^{\prime}} r) dr = \delta_{n,n^{\prime}} \frac{R^3}{2}[j_{l}(k_n R)]^2
	\label{eq:sph_ortho}
\end{equation}

\begin{figure*}[!t]
	\centerline{\includegraphics[width=11.0cm]{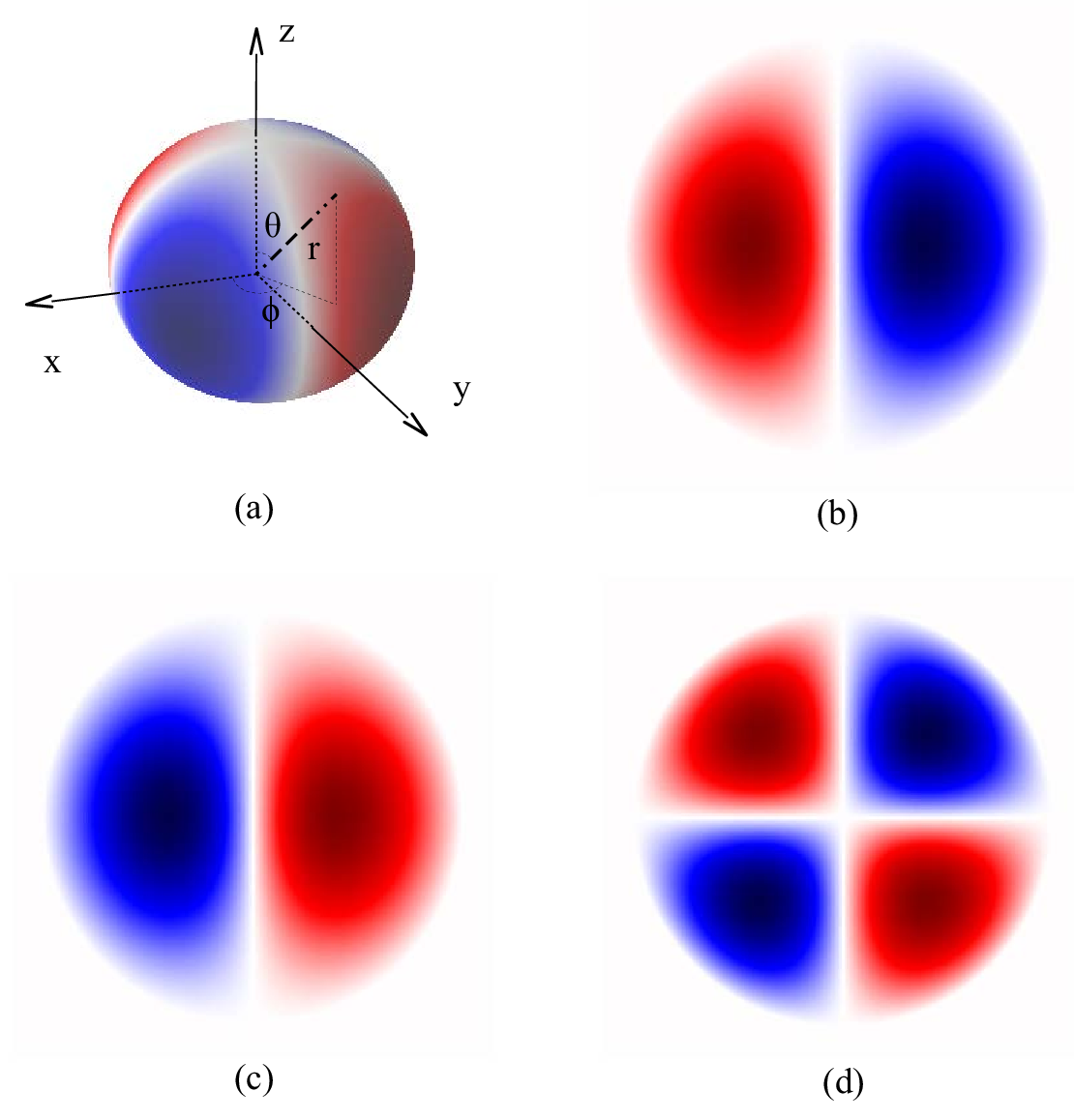}}
	\caption{The example of a spherical Fourier-Bessel basis with $l=1$, $m=2$ and $n=1$. (a) The overall 3D view of the basis function; (b) The cross-section perpendicular to the x-axis; (c) The cross-section perpendicular to the y-axis; (d) The cross-section perpendicular to the z-axis.}
	\label{fig_sfb}
\end{figure*}

In summary, the proposed 3D WMCG-CNN adopts a non-parameter-sharing strategy similar to the previous 2D WMCG-CNN\cite{zhao2023adaptive} study. The main differences are in two aspects. First, for the 3D WMCG-CNN, we develop a dedicated strategy to decompose the 3D transformation matrix of the $GL^+(3,\mathbb{R})$ continuous affine group. Second, the 3D WMCG-CNN uses a specially designed 3D filter basis (based on the spherical harmonics and the spherical Bessel function) to better extract features of volumetric data. The 3D WMCG convolution layer can be used to replace the standard non-scalar convolution layer in the residual CNN followed by a scalar convolution layer. As shown in the following experiments, for the convolution kernel size of $5\times 5\times 5$, this method only causes a minor increase in VRAM footprint and computational burden during the training phase. During the inference phase, the WMCG convolution layer has the same VRAM footprint and computational complexity as the standard convolution layer.

\section{Experiments}

\subsection{Datasets and experimental setup}
We validate our method on six datasets including four medical image datasets: BTCV\cite{landman2015miccai}, the Pancreas dataset\cite{roth2015deeporgan}, 3D-IRCADb-01\cite{soler20103d}, and the BraTS dataset\cite{baid2021rsna}; two seismic facies analysis datasets: the Parihaka dataset\cite{9686703} and the Netherlands F3 dataset\cite{alaudah2019machine}. All the tested networks are implemented with PyTorch. The training is performed on a GPU server with 8 A100 GPUs.

\subsubsection{BTCV dataset}
The BTCV dataset\cite{landman2015miccai} contains abdominal CT scans for 30 individuals. The CT scans were obtained with contrast enhancement in the portal venous phase. For each scan, 13 organs were annotated with the help of radiologists at Vanderbilt University Medical Center. Each scan contains 80 to 225 slices with a resolution of $512\times 512$ and a slice thickness of 1 to 6 mm.
The images are resampled into anisotropic voxel spacings of approximately 3.00 mm, 0.76 mm, and 0.76 mm for the three directions.

In the experiments with BTCV dataset and the other following datasets, the default training routine is based on nnUNet\cite{isensee2021nnu} but with a different optimizer setting according to \cite{roy2023mednext}. Neural networks are trained for $\num[group-separator={,}]{1000}$ epochs using the AdamW optimizer with an initial learning rate of 0.001, and a weight decay of 0.01. The input patch size is of $48 \times 192 \times 192$. The batch size is 2.

\subsubsection{BraTS dataset}

BraTS dataset\cite{baid2021rsna} consists of clinically-acquired multi-parametric MRI (mpMRI) scans of different brain tumors for 1251 patients from multiple data contributing institutions.
The mpMRI scans for each patients include
native (T1), post-contrast T1-weighted (T1Gd), T2-weighted (T2), and T2 Fluid Attenuated Inversion Recovery (T2-FLAIR) volumes.
The data were co-registered to the same anatomical template, interpolated to the same resolution (1 mm$^3$), and skull-stripped.
The annotations include the GD-enhancing tumor (ET), the peritumoral edematous/invaded tissue (ED), and the necrotic tumor core (NCR).

In the experiments with BraTS, the default training routine is based on nnUNet\cite{isensee2021nnu}, but with a different optimizer setting according to \cite{roy2023mednext}. The neural networks are trained for $\num[group-separator={,}]{30000}$ iterations using the ADAMW optimizer with an initial learning rate of 0.001 and weight decay of 0.01. The images are resampled to an isotropic voxel spacing of about 1.0 mm for the three directions. The input patch size is $96\times 96\times 96$. The batch size is 2.

\subsubsection{NIH Pancreas dataset}
The public NIH Pancreas dataset\cite{roth2015deeporgan} consists of contrast-enhanced abdominal CT scans from 82 patients. The pancreas was manually annotated on the CT images by an experienced radiologist. In the experiments, according to \cite{azad2024beyond}, 62 samples are used for training and 20 samples are used for testing.

In the experiments with the NIH Pancreas dataset, the default training routine is based on nnUNet\cite{isensee2021nnu}, but with a different optimizer setting according to \cite{roy2023mednext}. 
The neural networks are trained for $\num[group-separator={,}]{30000}$ iterations using the ADAMW optimizer with an initial learning rate of 0.001 and weight decay of 0.01. The input patch size is $96\times 96\times 96$. The batch size is 2.

\subsubsection{Liver vessel segmentation dataset}
The liver vessel segmentation dataset 3D-IRCADb-01\cite{soler20103d} consists of contrast-enhanced computed tomography scans for 20 patients. 15 cases of the dataset have hepatic tumors.
In the dataset, the hepatic veins, portal veins, and arteries are annotated.

In the experiments with 3D-IRCADb-01, the default training routine is based on nnUNet\cite{isensee2021nnu} but with a different optimizer setting according to \cite{roy2023mednext}. 
The neural networks are trained for $\num[group-separator={,}]{30000}$ iterations using the ADAMW optimizer with an initial learning rate of 0.001 and weight decay of 0.01. The images are resampled to anisotropy voxel spacings of approximately 1.6 mm, 0.74 mm, and 0.74 mm for the three directions.  The input patch size is $96\times 96\times 96$. The batch size is 2.

\subsubsection{Parihaka dataset}

The Parihaka dataset\cite{9686703} is the 3D seismic volume from the Taranaki Basin offshore New Zealand spanning an area of approximately $294$ $km^2$ with 590 inlines, 782 crosslines, and 1006 depth points. The volume data is annotated by expert geologists in Chevron with 6 facies: (1) Basement rocks, (2) Slope mudstone A, (3) Mass-transport complex, (4) Slope mudstone B, (5) Slope valley, and (6) Submarine canyon. The Slope mudstone facies make up 70\% of the total data. Similar to \cite{ORE2025105823}, the portion containing 20\% of the total inlines is used for testing. The facies classification (segmentation) on Parihaka dataset is usually performed in an inline-wise (2D) way as in \cite{9686703,ORE2025105823}. In our experiments with the Parihaka dataset, we adapt it for 3D segmentation.
For memory-saving in the 3D segmentation experiments, the volume is divided evenly into 75 sub-blocks. For a balance between classes, the facies are simplified into three classes: (1) Slope mudstone A, (2) Slope mudstone B, and (3) The rest.

The default training routine is based on nnUNet\cite{isensee2021nnu} but with a different optimizer setting according to \cite{roy2023mednext}. 
The neural networks are trained for $\num[group-separator={,}]{30000}$ iterations using the ADAMW optimizer with an initial learning rate of 0.001 and weight decay of 0.01. The input patch size is $96\times 96\times 96$. The batch size is 2.

\subsubsection{Netherlands F3 dataset}

The Netherlands F3 dataset\cite{alaudah2019machine} have 601 inlines, 901 crosslines, and 872 depth points for the F3 block located on the continental shelf of the North Sea in the Netherlands between two prominent graben structures.
The data is annotated by \cite{alaudah2019machine} with 6 facies: (1) Upper, (2) Middle, and (3) Lower North Sea group, (4) Rijnland/Chalk group, (5) Scruff group, and (6) Zechstein group. The North Sea groups make up 88\% of the total data. The facies classification (segmentation) on Netherlands F3 is usually performed in an inline-wise (2D) way as in \cite{alaudah2019machine,ORE2025105823}. In our experiments with the Netherlands F3 dataset, we adapt it for 3D segmentation.
For memory-saving in the 3D segmentation experiments, the volume is divided evenly into 36 sub-blocks
For a balance between classes, the facies are simplified into four classes: (1) Upper, (2) Middle, (3) Lower North Sea group, and (4) The rest.

The default training routine is based on nnUNet\cite{isensee2021nnu} but with a different optimizer setting according to \cite{roy2023mednext}. 
The neural networks are trained for $\num[group-separator={,}]{30000}$ iterations using the ADAMW optimizer with an initial learning rate of 0.001 and weight decay of 0.01. The input patch size is $96\times 96\times 96$. The batch size is 2.

\subsection{Ablation experiments}
The ablation experiments are performed on BTCV dataset. To save time, the number of training epochs is set to 200. Within this number of epochs one already sees the tendency and thus it is a good strategy to do ablation while not consuming too much time. The mean Dice Similarity Coefficient (DSC) and the mean Surface Dice Coeffcient (SDC) are reported with standard deviation $\sigma$, where $\sigma = \sqrt{[\Sigma (x_i -\mu)^2]/N}$ with $N$ the population size, $x_i$ the $i$-th value from the population, $\mu$ the mean value.

For implementation of our methods, we adopt the filter decomposition as equation \eqref{eq:matrixdecomp}. We choose MedNeXt-S-k5 from \cite{roy2023mednext} as a base model to build a corresponding WMCG-CNN model. The default setting of the WMCG-CNN model is denoted as "WMCG-sFB-k5-nb27-shear0.25$\pi$", where the spherical Fourier-Bessel bases are augmented with uniform random circular shift, random rotation, random shear transformation with shear angle in the range of $[-0.25\pi,0.25\pi)$, isotropic random scaling in the range of $[1,2)$. The scaling coefficients for the three directions have the same value, while the other augmentation parameters are all sampled independently. 
To justify the effectiveness of the default setting, we test the cases of using various ranges of shear angle, using anisotropic scaling coefficients, using only the spherical harmonics\cite{weiler20183d,cesa2021program}, and using kernel reparameterization\cite{lee2023scaling}.

The suffixes in the model names are used to denote the models with different settings. "k$n$" means the filter size of $n\times n\times n$. "nb$n$" means $n$ bases used per filter. "sph" means the spherical harmonic bases as equation \ref{eq:sph}. "sFB" refers to the spherical Fourier-Bessel bases as equation \ref{eq:sFB}. "diffscale" means that the scaling coefficients for three directions are sampled independently.

The results are shown in Table \ref{tababaltion}. In addition to DSC and SDC, we report the number of trainable parameters in million ($10^6$), Params(M); the number of multiply-accumulate operations in giga ($10^9$), MACs(G)\footnote{In all the following experiments, Params and MACs are calculated assuming that the inputs are images from the BTCV dataset. }; Significance of the difference, SD, the p-value of the paired Wilcoxon signed-rank test on the comparison with the target methods (marked with the asterisk symbol "*").  We see that an appropriate shear angle is beneficial for performance. 
Large shear angles can be detrimental to performance, possibly because large shear angles can lead to a loss of filter information in discrete implementations. 
Anisotropic scaling is shown to degrade segmentation performance. The possible reason is that the isotropic scaling units (i.e., the pooling method) are more common in deep neural networks. Therefore, in all the following experiments, we use isotropic scaling and the default range of shear angles.

\begin{table*}
	\caption{Ablation experimental results on the BTCV dataset.}
	\label{table}
	\setlength{\tabcolsep}{3pt}
	\centering
	\centerline{\resizebox{12.6cm}{!}{ 
			\begin{threeparttable}[b]
				\begin{tabular}{cccccc}
					\toprule
					Model& 
					Params (M)& 
					MACs (G)&
					DSC &
					SDC & SD (w.r.t. DSC+SDC)\\
					\midrule
					WMCG-sFB-k5-nb27-shear0.00 & 
					5.56  & 
					141.94 &
					81.74$\pm$2.78 &
					75.78$\pm$2.36 & 0.313(*)\\ 
					WMCG-sFB-k5-nb27-shear0.10$\pi$ &
					5.56  & 
					141.94 &
					\textbf{82.15$\pm$2.49} &
					\textbf{76.42$\pm$1.94} & - \\
					WMCG-sFB-k5-nb27-shear0.25$\pi$&
					5.56  & 
					141.94 &
					82.00$\pm$2.04 &
					76.12$\pm$1.72& *\\
					WMCG-sFB-k5-nb27-shear0.40$\pi$ &
					5.56  & 
					141.94 &
					81.56$\pm$2.44 &
					75.87$\pm$2.12 & 0.156(*)\\
					WMCG-sFB-k5-nb27-diffscale &
					5.56 & 
					141.94 &
					81.60$\pm$2.58 &
					75.81$\pm$2.02 & 0.156(*) \\
					WMCG-sph-k5-nb27-shear0.25$\pi$&
					5.56  & 
					141.94 &
					81.69$\pm$2.69 &
					75.99$\pm$2.06 & 0.406(*)\\
					WMCG-sFB-k5-nb27-scale0.00 &
					5.56  & 
					141.94 &
					81.87$\pm$2.54 &
					76.14$\pm$1.94 &  0.594(*)  \\
					WMCG-sFB-k5-nb27-scale0.50 &
					5.56  & 
					141.94 &
					82.09$\pm$2.20 &
					76.31$\pm$2.04 & - \\
					\bottomrule
				\end{tabular}
	\end{threeparttable}}}
	\label{tababaltion}
\end{table*}

\subsection{Affine group equivariance tests}
Using MedNeXt as base model, we compare the proposed convolutional layer with the standard CNN layer and the state-of-the-art 3D group equivariant convolutional layer: SE(3)\cite{weiler20183d} and PDO\cite{he2023neural}.
The model with SE(3) or PDO is constructed by replacing all hidden $5\times 5\times 5$-kernel convolutional layers (except for the downsampling and upsampling layers) with the corresponding SE(3)\cite{weiler20183d} or PDO\cite{he2023neural} G-CNN layers with the same kernel size. The models with WMCG-sFB are constructed by replacing all hidden $5\times 5\times 5$-kernel convolutional layers (except for the downsampling and upsampling layers) with the proposed WMCG-sFB G-CNN layers using the first 27 bases. 
To test affine group equivariance of the proposed method, we use the BraTS dataset with isotropy voxel spacing. Another reason we choose BraTS dataset is because its resolutions for three directions are closer to each other than other datasets, which allows random affine transformations with less information loss. We augment the loaded images with random affine transformation with the shear range $(-0.5\pi,0.5\pi)$, the scaling range $[1.0,2.0)$ and rotation angle range $[-\pi,\pi)$. For stability, the neural networks are trained with a learning rate of 1e-4.

The results on group equivariance are shown in Table \ref{tab-geerrors}. The significance level (SD) above $0.05$ is marked with bold font. The proposed convolutional layers exhibit superior affine group equivariance.

\begin{table*}
	\caption{The average segmentation performance on affine group equivariance on the BraTS dataset.}
	\setlength{\tabcolsep}{3pt}
	\centering
	\centerline{\resizebox{10.5cm}{!}{ 
			\begin{threeparttable}[b]
				\begin{tabular}{cccccc}
					\toprule
					\multirow{2}{*}{Base model} & \multicolumn{2}{c}{MedNeXt-S-k5}                              & \multicolumn{2}{c}{MedNeXt-L-k5}                              & \multirow{2}{*}{\minitab[c]{SD (w.r.t. \\DSC+SDC)}} \\ \cmidrule(lr){2-5}
					& \multicolumn{1}{c}{DSC}                 & SDC                 & \multicolumn{1}{c}{DSC}                 & SDC            &            \\ \midrule
					CNN        & \multicolumn{1}{c}{69.71±1.20}          & 56.17±1.63          & \multicolumn{1}{c}{68.67±1.16}          & 54.24±0.88          &    \textbf{ 0.001(*) }     \\ 
					SE(3)\cite{weiler20183d}       & \multicolumn{1}{c}{68.30±2.40}          & 54.69±1.94          & \multicolumn{1}{c}{68.51±2.96}          & 56.29±1.75          &    \textbf{ 0.001(*) }   \\ 
					PDO\cite{he2023neural}        & \multicolumn{1}{c}{70.21±3.02}          & 57.41±3.42          & \multicolumn{1}{c}{68.71±0.29}          & 54.31±1.08          &     \textbf{0.042(*)}    \\ 
					WMCG-sFB        & \multicolumn{1}{c}{\textbf{71.04±1.07}} & \textbf{58.64±1.12} & \multicolumn{1}{c}{\textbf{70.81±0.95}} & \textbf{58.71±1.00} &    *     \\ \bottomrule
				\end{tabular}
	\end{threeparttable}}}
	\label{tab-geerrors}
\end{table*}

\subsection{Image segmentation experiments with different sample sizes}
We test the proposed methods on four public 3D image segmentation datasets.
The neural networks are first trained with a learning rate of 1e-3. If NaN (the training loss "not a number") errors occur, the neural networks are retrained from scratch for all split folds with a learning rate of 1e-4. The best performance for each split fold is reported.

To test the data efficiency of the proposed convolutional layers, we design small sample medical image segmentation experiments. For the 2-shot and 10-shot experiments, 2 and 10 random samples are selected as the training set, and the rest samples are used for test. The experiments for each model are repeated for eight times. The number of samples in the test sets for the BraTS dataset is limited to 251.

The experimental results in Table \ref{tab-fewshot} show that the proposed convolutional layers are superior to plain CNN and SE(3)\cite{weiler20183d} on average for both 2-shot and 10-shot learning. The advantage of our methods in 10-shot learning is higher than that in 2-shot learning.
Table \ref{tab-fewshot} also shows the results for different filter augmentation strategies. Shear0.00, Scale0.00, and Rotation0.00 represent no shear augmentation, no scale augmentation, and no rotation augmentation, respectively. Affine0.00 represents no shear, scale, and rotation augmentation. The results show that, on average, insufficient augmentation leads to poor performance of DSC+SDC.

The experimental results on non-small-sample tests are shown in Table \ref{tab-fivefolds}, where the training data splitting follows the default setting of nnUNet\cite{isensee2021nnu}. 
To verify the leading position of the adopted ADCNN MedNeXt, the other recent state-of-the-art networks are also tested including nnUNet\cite{isensee2021nnu}, nnUNet-Res-Enc\cite{isensee2024nnu}, UNETR\cite{hatamizadeh2022unetr}, UNETR++\cite{shaker2024unetr++}, UXNet\cite{lee20223d}, UMamba\cite{U-Mamba}, SegMamba\cite{Xin_SegMamba_MICCAI2024}, VSmTrans\cite{liu2024vsmtrans}, and STUNet\cite{huang2023stu}. These networks are trained with their respective optimal default settings using the same number of iterations as MedNeXt\cite{roy2023mednext}.
	In Table \ref{tab-extend-param}, the Training VRAM Footprint (MB), Inference Time (s) per volume, Params (M), and MACs (G) of the neural network models for the BTCV dataset are reported. 
	We see that, on average, our methods achieve a competitive or better performance than both the standard CNN layers and the existing G-CNN layers\cite{weiler20183d} on the four different datasets. 
	In addition, the results show that MedNeXt is the state-of-the-art deep CNN for the dense volumetric image segmentation tasks. The different versions of MedNeXt show a good balance between performance and costs of computation and storage. The proposed methods show a high compatibility with MedNeXt. Our methods are competitive or superior to other methods on the six public datasets.


\begin{table*}
	\caption{The average small-sample test results on the six different datasets.}
	\label{table}
	\setlength{\tabcolsep}{3pt}
	\centering
	\centerline{\resizebox{16.2cm}{!}{ 
			\begin{threeparttable}[b]
				\begin{tabular}{cccccccccc}
					\toprule
					2 shots                   &     & BTCV                & NIH Pancreas        & 3D-IRCADb            & BraTS               & Parihaka             & Netherlands F3       & Average        & SD (w.r.t. DSC+SDC) \\ \midrule
					\multirow{2}{*}{CNN}      & DSC & 40.82±4.55          & 53.54±5.93          & 51.41±15.05          & 70.47±6.79          & 28.46±11.85          & 44.14±11.22          & 48.14          & \multirow{2}{*}{\textbf{0.006(*)}}  \\ 
					& SDC & 33.99±3.87          & 41.73±4.78          & 47.72±15.14          & 63.90±7.01          & 19.31±8.66           & 33.27±6.78           & 39.99          &                     \\ \midrule
					\multirow{2}{*}{SE(3)\cite{weiler20183d}}    & DSC & \textbf{45.14±5.12} & 48.81±7.10          & 51.46±14.26          & 69.69±7.98          & 30.49±13.18          & 44.32±12.09          & 48.32          & \multirow{2}{*}{0.236(*)}  \\ 
					& SDC & \textbf{38.02±4.50} & 37.76±5.90          & 47.31±14.74          & 63.01±8.00          & 20.53±9.89           & 32.29±6.04           & 39.82          &                     \\ \midrule
					\multirow{2}{*}{PDO\cite{he2023neural}}      & DSC & 42.44±4.98          & 44.52±8.95          & 47.88±15.00          & 69.33±9.79          & \textbf{32.16±11.37} & 43.55±11.09          & 46.65          & \multirow{2}{*}{\textbf{0.001(*)}}  \\ 
					& SDC & 35.21±4.47          & 34.51±7.21          & 44.28±15.28          & 63.06±9.70          & \textbf{21.23±8.80}  & 32.60±5.91           & 38.48          &                     \\ \midrule
					\multirow{2}{*}{WMCG-sFB} & DSC & 42.80±5.37          & 52.52±6.47          & 51.96±14.73          & \textbf{71.16±7.65} & 29.01±12.26          & \textbf{44.82±13.18} & 48.71          & \multirow{2}{*}{*}  \\ 
					& SDC & 35.85±4.78          & 40.97±5.71          & 47.23±15.71          & \textbf{65.49±7.60} & 20.01±9.05           & \textbf{33.73±7.61}  & \textbf{40.55} &                     \\ \midrule
					\multirow{2}{*}{Shear0.00}  & DSC & 42.02±5.39          & 53.17±7.04          & 52.55±13.97          & 70.94±6.46          & 30.28±11.73          & 43.35±11.23          & 48.72          & \multirow{2}{*}{0.105(*)}  \\ 
					& SDC & 35.02±4.76          & 41.25±5.92          & 47.74±15.21          & 64.92±7.06          & 20.23±8.52           & 32.36±6.66           & 40.25          &                     \\ \midrule
					\multirow{2}{*}{Scale0.00}  & DSC & 41.71±5.34          & 53.17±5.80          & 51.86±14.39          & 70.61±7.67          & 30.43±11.96          & 44.57±12.62          & 48.73          & \multirow{2}{*}{0.073(*)}  \\ 
					& SDC & 34.94±4.60          & 41.41±5.07          & 47.34±15.40          & 64.75±7.67          & 20.53±8.58           & 33.29±7.20           & 40.37          &                     \\ \midrule
					\multirow{2}{*}{Rotation0.00}    & DSC & 42.66±4.48          & 53.09±6.44          & 52.01±14.72          & 70.61±7.34          & 28.29±10.70          & 43.91±12.76          & 48.43          & \multirow{2}{*}{0.266(*)}  \\ 
					& SDC & 35.93±3.94          & 41.44±5.47          & 47.62±15.22          & 64.66±7.29          & 19.61±8.51           & 33.02±7.30           & 40.38          &                     \\ \midrule
					\multirow{2}{*}{Affine0.00} & DSC & 40.66±5.30          & \textbf{53.72±5.92} & \textbf{52.75±13.70} & 70.99±6.59          & 30.31±12.66          & 44.08±12.25          & \textbf{48.75} & \multirow{2}{*}{0.104(*)}  \\ 
					& SDC & 33.83±4.62          & \textbf{41.93±4.88} & \textbf{48.32±14.21} & 64.48±7.87          & 20.49±9.23           & 32.81±7.14           & 40.31          &                     \\ \midrule
					10 shots                  &     & BTCV                & NIH Pancreas        & 3D-IRCADb            & BraTS               & Parihaka             & Netherlands F3       & Average        & SD (w.r.t. DSC+SDC) \\ \midrule
					\multirow{2}{*}{CNN}      & DSC & 73.82±6.08          & 76.87±1.88          & 65.53±3.83           & 80.79±2.33          & 55.85±6.31           & 73.01±7.62           & 70.98          & \multirow{2}{*}{\textbf{0.000(**)}} \\ 
					& SDC & 66.09±7.22          & 63.88±1.94          & 62.31±2.46           & 77.53±2.98          & 39.03±3.34           & 53.35±3.45           & 60.37          &                     \\ \midrule
					\multirow{2}{*}{SE(3)\cite{weiler20183d}}    & DSC & 76.51±2.32          & 76.12±1.94          & 65.63±3.00           & 81.44±1.63          & 54.39±5.80           & 73.14±7.88           & 71.20          & \multirow{2}{*}{\textbf{0.000(**)}} \\ 
					& SDC & 68.79±4.05          & 62.78±1.59          & 61.81±1.91           & 78.28±2.20          & 37.86±2.77           & 53.21±3.42           & 60.45          &                     \\ \midrule
					\multirow{2}{*}{PDO\cite{he2023neural}}      & DSC & 75.54±1.43          & 74.26±2.67          & 64.50±2.54           & 80.92±1.55          & 55.79±4.73           & 73.19±7.88           & 70.70          & \multirow{2}{*}{\textbf{0.000(**)}} \\ 
					& SDC & 68.24±1.37          & 60.85±2.36          & 60.61±1.78           & 77.97±2.09          & 38.01±2.33           & 53.17±3.67           & 59.81          &                     \\ \midrule
					\multirow{2}{*}{WMCG-sFB} & DSC & \textbf{77.88±1.06} & \textbf{78.16±1.67} & \textbf{66.47±3.62}  & \textbf{82.17±1.58} & \textbf{56.07±6.19}  & 73.63±8.73           & \textbf{72.40} & \multirow{2}{*}{**} \\ 
					& SDC & \textbf{71.50±1.29} & \textbf{65.16±1.51} & \textbf{62.69±2.75}  & 79.45±2.07          & \textbf{39.27±2.60}  & 53.82±4.15           & \textbf{61.98} &                     \\ \midrule
					\multirow{2}{*}{Shear0.00}  & DSC & 77.60±1.11          & 77.91±1.77          & 65.94±3.31           & 81.80±1.62          & 55.18±5.51           & 73.87±8.68           & 72.05          & \multirow{2}{*}{\textbf{0.003(**)}} \\ 
					& SDC & 70.98±1.36          & 65.14±1.60          & 61.94±2.70           & 79.07±2.26          & 38.56±2.37           & 53.81±3.89           & 61.58          &                     \\ \midrule
					\multirow{2}{*}{Scale0.00}  & DSC & 77.04±0.98          & 78.08±1.82          & 66.08±3.54           & 80.99±1.80          & 54.59±6.15           & \textbf{74.40±7.97}  & 71.86          & \multirow{2}{*}{\textbf{0.002(**)}} \\ 
					& SDC & 70.38±1.29          & 65.10±1.61          & 62.34±2.59           & 78.17±2.48          & 38.37±2.67           & \textbf{54.22±3.80}  & 61.43          &                     \\ \midrule
					\multirow{2}{*}{Rotation0.00}    & DSC & 77.37±1.16          & 77.70±1.99          & 66.26±2.91           & 81.91±1.60          & 55.52±6.77           & 73.90±8.53           & 72.11          & \multirow{2}{*}{\textbf{0.014(**)}} \\ 
					& SDC & 70.93±1.35          & 64.65±1.73          & 62.49±2.23           & 79.23±2.12          & 38.96±2.99           & 53.93±3.93           & 61.70          &                     \\ \midrule
					\multirow{2}{*}{Affine0.00} & DSC & 77.47±0.89          & 77.88±1.93          & 65.82±3.13           & 82.04±1.63          & 55.56±5.57           & 73.10±8.34           & 71.98          & \multirow{2}{*}{\textbf{0.017(**)}} \\ 
					& SDC & 70.87±1.14          & 64.76±1.88          & 62.36±2.20           & \textbf{79.53±2.04} & 38.99±2.68           & 53.92±4.23           & 61.74          &                     \\ \bottomrule
				\end{tabular}
	\end{threeparttable}}}
	\label{tab-fewshot}
\end{table*}

\begin{table*}
	\caption{The comparison of segmentation performance of different models on the six different datasets with normal default data splits. The rows in light gray and gray show the results for the transformer-based models and Mamba-based models, respectively.}
	\label{table}
	\setlength{\tabcolsep}{3pt}
	\centering
	\centerline{\resizebox{18.0cm}{!}{ 
			\begin{threeparttable}[b]
				\begin{tabular}{ccccccccccc}
					\toprule
					Model                                                              & Base Model                                                       & \multicolumn{1}{l}{Metrics} & BTCV                & NIH Pancreas        & 3D-IRCADb           & BraTS               & Parihaka            & Netherlands F3      & Average        & SD (w.r.t. DSC+SDC)          \\ \midrule
					\multirow{2}{*}{nnUNet\cite{isensee2021nnu}}      &                                             & DSC                          & 83.00±2.73          & 85.23±0.96          & 63.73±9.67          & 90.28±0.70          & 60.79±3.46          & 70.45±2.41          & 75.58          & \multirow{2}{*}{\textbf{0.000(**)}} \\ 
					& \multicolumn{1}{l}{}                                            & SDC                          & 77.84±2.35          & 73.68±1.99          & 61.05±5.69          & 88.98±0.80          & 37.29±2.63          & 49.69±2.69          & 64.75          &                              \\ \midrule
					\multirow{2}{*}{nnUNet-Res-Enc-L\cite{isensee2024nnu}}                                  &                                                                  & DSC                          & 83.37±2.76          & 85.64±1.10          & 63.93±5.66          & \textbf{90.70±0.70} & 66.71±2.49          & 72.79±3.11          & 77.19          & \multirow{2}{*}{\textbf{0.003(**)}} \\ 
					&                                                                  & SDC                          & 78.70±2.19          & 74.80±1.98          & 61.14±2.04          & 89.41±0.77          & 45.97±3.02          & 53.34±3.58          & 67.23          &                              \\ \midrule
					\multirow{2}{*}{UXNET\cite{lee20223d}}            &                                                & DSC                          & 79.38±2.29          & 82.16±1.35          & 61.80±8.18          & 89.92±0.66          & 59.57±1.79          & 73.47±4.62          & 74.38          & \multirow{2}{*}{\textbf{0.000(**)}} \\ 
					&                                                                  & SDC                          & 72.42±1.90          & 69.35±1.84          & 57.45±5.20          & 88.44±0.78          & 36.96±2.08          & 51.21±4.32          & 62.64          &                              \\ \midrule
					\multirow{2}{*}{STUNet-L\cite{huang2023stu}}                                          &                                                      & DSC                          & 83.12±2.65          & 84.89±1.08          & 63.38±7.16          & 90.11±0.81          & 61.08±2.70          & 72.27±5.36          & 75.81          & \multirow{2}{*}{\textbf{0.000(**)}} \\ 
					&                                                                  & SDC                          & 78.32±2.19          & 73.47±2.09          & 60.38±3.66          & 88.79±0.84          & 37.46±2.76          & 52.44±3.49          & 65.14          &                              \\ \midrule
					\rowcolor{deflightgray}	 &                                    & DSC                          & 71.16±2.93          & 76.06±1.59          & 58.74±4.38          & 87.87±0.71          & 49.19±6.04          & 68.75±1.98          & 68.63          &  \\ 
					\rowcolor{deflightgray}\multirow{-2}{*}{UNETR\cite{hatamizadeh2022unetr}}&                                                             & SDC                          & 61.89±2.69          & 59.78±2.30          & 53.31±2.14          & 85.14±0.77          & 27.95±4.78          & 49.08±3.68          & 56.19          &           \multirow{-2}{*}{\textbf{0.000(**)}}                   \\ \midrule
					\rowcolor{deflightgray}  &                                          & DSC                          & 81.06±1.66          & 85.10±1.41          & 61.14±7.54          & 88.80±0.92          & 57.36±3.64          & 70.46±4.63          & 73.99          &  \\ 
					\rowcolor{deflightgray}\multirow{-2}{*}{UNETR++\cite{shaker2022unetr++}}&                                                                  & SDC                          & 74.53±1.33          & 73.58±2.05          & 57.73±3.85          & 86.82±0.76          & 36.71±3.63          & 52.66±3.35          & 63.67          &             \multirow{-2}{*}{\textbf{0.000(**)}}                 \\ \midrule
					\rowcolor{deflightgray}	                                          &                                            & DSC                          & 80.53±2.21          & 84.66±0.97          & 60.71±5.80          & 90.17±0.65          & 65.66±3.39          & 73.06±3.85          & 75.80          & \\ 
					\rowcolor{deflightgray}\multirow{-2}{*}{VSmTrans\cite{liu2024vsmtrans}}&                                                                      & SDC                          & 74.20±1.55          & 73.02±1.77          & 56.70±1.79          & 88.62±0.80          & 41.08±3.32          & 51.34±3.48          & 64.16          &            \multirow{-2}{*}{\textbf{0.000(**)}}                   \\ \midrule
					\rowcolor{defgray}	                                            &                                                  & DSC                          & 83.52±2.52          & 83.72±1.37          & 63.99±5.81          & 88.41±1.28          & 60.20±7.23          & 74.85±2.72          & 75.78          &  \\ 
					\rowcolor{defgray}\multirow{-2}{*}{UMamba\cite{U-Mamba}}	&                                                                  & SDC                          & 78.77±1.86          & 70.95±2.46          & 60.73±2.95          & 86.62±1.24          & 35.33±4.57          & 51.84±1.46          & 64.04          &              \multirow{-2}{*}{\textbf{0.000(**)}}                \\ \midrule
					\rowcolor{defgray}	                                          &                                                      & DSC                          & 80.98±2.54          & 84.25±1.32          & 63.27±6.76          & 90.04±0.76          & 56.79±2.50          & 73.90±4.34          & 74.87          &  \\ 
					\rowcolor{defgray}\multirow{-2}{*}{SegMamba\cite{Xin_SegMamba_MICCAI2024}}	&                                                                   & SDC                          & 74.86±2.05          & 72.05±2.18          & 59.26±4.17          & 88.42±0.86          & 32.09±1.25          & 49.47±3.23          & 62.69          &             \multirow{-2}{*}{\textbf{0.000(**)}}                 \\ \midrule
					\multirow{2}{*}{CNN}                                               & \multirow{8}{*}{MedNeXt-S\cite{roy2023mednext}} & DSC                          & 82.53±2.33          & 85.64±1.26          & 64.08±6.20          & 90.31±0.76          & 67.02±2.66          & 77.39±3.06          & 77.83          & \multirow{2}{*}{\textbf{0.019(*)}}           \\ 
					&                                                                  & SDC                          & 77.01±2.21          & 74.84±2.21          & 61.23±3.31          & 88.96±0.81          & 47.45±1.98          & 55.49±2.89          & 67.50          &                              \\ \cmidrule(lr){1-1} \cmidrule(lr){3-11}
					\multirow{2}{*}{SE(3)\cite{weiler20183d}}                                             &                                                                  & DSC                          & 82.48±2.34          & 85.60±1.01          & 63.87±8.20          & 90.68±0.68          & 67.88±2.07          & 77.90±2.37          & 78.07          & \multirow{2}{*}{0.099(*)}           \\ 
					&                                                                  & SDC                          & 76.97±2.34          & 74.53±2.05          & 61.24±4.27          & \textbf{89.45±1.00} & 46.80±3.58          & 55.36±3.13          & 67.39          &                              \\ \cmidrule(lr){1-1} \cmidrule(lr){3-11}
					\multirow{2}{*}{PDO\cite{he2023neural}}                                               &                                                                  & DSC                          & 81.40±2.45          & 85.20±0.88          & 61.79±6.39          & 90.37±0.69          & 67.68±1.31          & 74.89±3.27          & 76.89          & \multirow{2}{*}{\textbf{0.000(*)}}           \\ 
					&                                                                  & SDC                          & 75.42±1.95          & 73.47±1.58          & 58.77±3.72          & 89.10±0.89          & 45.37±2.63          & 54.18±3.00          & 66.05          &                              \\ \cmidrule(lr){1-1} \cmidrule(lr){3-11}
					\multirow{2}{*}{WMCG-sFB}                                          &                                                                  & DSC                          & 82.81±2.41          & 85.76±1.19          & 64.54±6.61          & 90.39±0.50          & 68.31±1.80          & 77.95±3.39          & 78.29          & \multirow{2}{*}{*}           \\ 
					&                                                                  & SDC                          & 77.46±2.20          & 74.92±2.20          & \textbf{62.18±3.70} & 89.06±0.71          & 46.71±2.67          & 56.01±3.11          & 67.72          &                              \\ \midrule
					\multirow{2}{*}{CNN}                                               & \multirow{8}{*}{MedNeXt-L\cite{roy2023mednext}} & DSC                          & 82.56±2.85          & 86.04±0.99          & 63.53±7.19          & 89.97±0.59          & 66.63±3.43          & 77.29±3.53          & 77.67          & \multirow{2}{*}{\textbf{0.000(**)}} \\ 
					&                                                                  & SDC                          & 77.07±2.91          & 75.33±2.08          & 61.09±4.30          & 88.57±0.76          & 45.44±2.86          & 55.12±3.42          & 67.10          &                              \\ \cmidrule(lr){1-1} \cmidrule(lr){3-11}
					\multirow{2}{*}{SE(3)\cite{weiler20183d}}                                             &                                                                  & DSC                          & 82.91±2.34          & 85.93±0.89          & 64.63±6.84          & 90.37±0.60          & 67.45±2.79          & 76.28±3.07          & 77.93          & \multirow{2}{*}{\textbf{0.008(**)}} \\ 
					&                                                                  & SDC                          & 77.95±2.14          & 74.80±1.72          & 61.69±3.38          & 89.12±0.89          & 46.99±3.28          & 54.23±3.23          & 67.46          &                              \\ \cmidrule(lr){1-1} \cmidrule(lr){3-11}
					\multirow{2}{*}{PDO\cite{he2023neural}}                                               &                                                                  & DSC                          & 82.04±2.56          & 84.98±1.20          & 62.62±7.07          & 90.51±0.69          & 66.04±3.19          & 76.56±1.82          & 77.12          & \multirow{2}{*}{\textbf{0.000(**)}} \\ 
					&                                                                  & SDC                          & 76.36±1.96          & 73.03±1.93          & 60.07±3.87          & 89.42±0.85          & 42.34±2.60          & 54.53±2.59          & 65.96          &                              \\ \cmidrule(lr){1-1} \cmidrule(lr){3-11}
					\multirow{2}{*}{WMCG-sFB}                                          &                                                                  & DSC                          & \textbf{83.52±2.64} & \textbf{86.09±1.04} & \textbf{64.66±6.84} & 90.45±0.62          & \textbf{68.06±2.14} & \textbf{78.37±2.77} & \textbf{78.53} & \multirow{2}{*}{\textbf{**}} \\ 
					&                                                                  & SDC                          & \textbf{78.82±2.46} & \textbf{75.34±2.10} & 61.96±3.07          & 89.16±0.94          & \textbf{47.83±3.30} & \textbf{56.24±2.83} & \textbf{68.23} &                              \\ \bottomrule
				\end{tabular}
	\end{threeparttable}}}
	\label{tab-fivefolds}
\end{table*}

\subsection{Training stability results}
Table \ref{tab-nanerrors} summarizes the frequency of NaN training loss errors that occurred during the experiments with MedNeXt-k5 networks for Table \ref{tab-fivefolds} and Table \ref{tab-fewshot}. It proves the superior training stability of the proposed method. It confirms that higher affine equivariance correlates with higher training stability.

\begin{table*}[!htbp]
	\caption{The frequency of NaN training loss errors that happened during the experiments with MedNeXt-k5 networks for Table \ref{tab-fivefolds} and Table \ref{tab-fewshot}.}
	\setlength{\tabcolsep}{3pt}
	\centering
	\centerline{\resizebox{12.6cm}{!}{ 
			\begin{threeparttable}[b]
				\begin{tabular}{ccccccccc}
					\toprule
					Model & BTCV & NIH Pancreas & 3D-IRCADb & BraTS &  Parihaka            & Netherlands F3      & Sum  & SD      \\ \midrule
					CNN & 4    & 0            & 0      & 10    & 0 & 0 & 14   &  \textbf{0.043(*)}    \\ 
					SE(3)\cite{weiler20183d}    & 0    & 0            & 0      & 5   & 0 & 0  & 5   &   0.159(*)    \\ 
					PDO\cite{he2023neural}    & 0    & 5            & 4      & 13  & 5 & 1  & 28  &   \textbf{0.008(*) }    \\ 
					WMCG-sFB      & \textbf{0}    & \textbf{0 }           & \textbf{0}      & \textbf{1}  & \textbf{0} & \textbf{0}  & \textbf{1} & *  \\ \bottomrule
				\end{tabular}
	\end{threeparttable}}}
	\label{tab-nanerrors}
\end{table*}

\section{Discussion and conclusion}
In this work, we propose an efficient non-parameter-sharing 3D affine G-CNN by weighted aggregation of Monte Carlo augmented spherical Fourier-Bessel bases. The proposed method is superior to the state-of-the-art 3D G-CNN in 3D affine group equivariance for the dense prediction task, the volumetric image segmentation. 
Accordingly, superior training stability and average data efficiency over conventional CNNs are observed in the experiments on four different volumetric medical image datasets and two seismic datasets. Our method is flexible and pluggable for integration into various existing deep CNN models or ADCNNs. 

Compared with the existing methods on 3D G-CNNs, our approach introduces a set of 3D filter bases with high expressiveness based on the spherical harmonics and the spherical Bessel functions, which combines radial orthogonality and angular orthogonality. Using these novel bases and the non-parameter-sharing weighting strategy, we build a powerful, flexible, and efficient affine equivariant convolutional operator.

\begin{figure*}[!t]
	\centerline{\includegraphics[width=14.0cm]{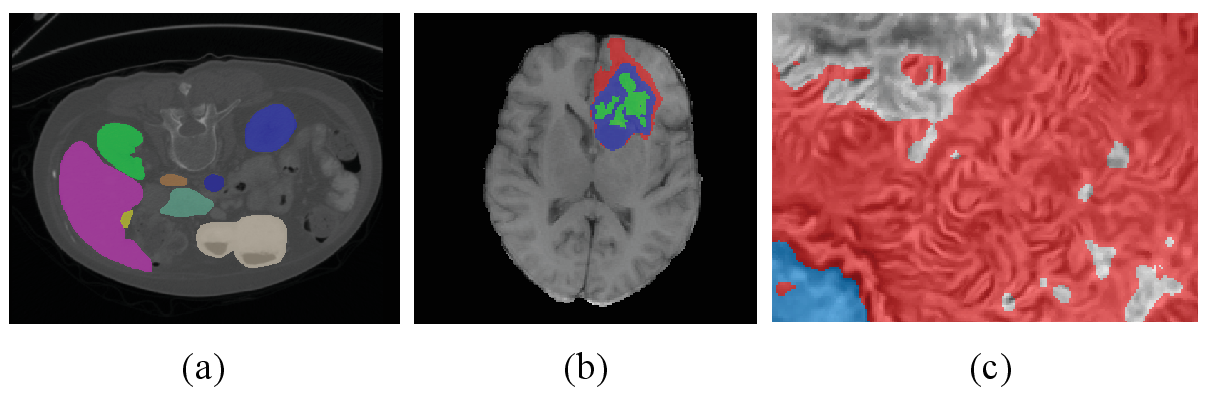}}
	\caption{The examples of axial slices with colored segmentation labels for different datasets. (a) BTCV; (b) BRaTS; (c) Netherlands F3}.
	\label{fig_btcv_brats}
\end{figure*}

It should be noted that for experiments on the BraTS dataset with the common training routine in Table \ref{tab-fivefolds}, our affinely equivariant CNNs fail to outperform the rotationally equivariant CNN (SE(3)\cite{weiler20183d}) in segmentation accuracy. This is likely due to the fact that brain images (as shown in Fig. \ref{fig_btcv_brats}) in real life mainly have global rigid transformations such as rotation, whereas abdomain images are rich in local elastic deformations and the local structures often undergo local affine transformations. 
Meanwhile, in another experiment on BraTS dataset against global affine transformation as in Table \ref{tab-geerrors}, our methods show superior performance to SE(3)\cite{weiler20183d}. This indicates that our methods are better at dealing with some complex transformation like affine transformation. In contrast, SE(3) is dedicated to global rotation equivariance and usually performs poorly in dealing with affine transformations.

Another limitation of our methods is the increase in GPU’s VRAM (Video Random Access Memory) usage due to the adaptive fusion (i.e., the weighted sum) of filters during the training phase. For the kernel sizes such as $5\times 5 \times 5$, as shown in Table \ref{tab-extend-param}, the increase in memory usage is tiny. Clearly, for 3D kernels, the memory usage will increase drastically along with the kernel size. To mitigate this problem, a safe solution is to decompose the large kernel into the convolutions between multiple small kernels. Since most state-of-the-art neural networks use small kernels, and the extension to large kernels is not the focus of this study, we leave this for future work.

Our methods involve the Monte Carlo sampling of the 3D transformation matrices, which shows a great flexibility in handling the high-dimensional space of transformation parameters. But the Monte Carlo sampling has a slow converging speed compared to the uniform sampling. 
The slow convergence speed of Monte Carlo sampling results in a high sampling variance for cases with a small number of samples, and may cause information loss and instability when a single filter basis function is used.	However, in practice, multiple bases are usually used, which increases the diversity of the filters. Moreover, in each bottleneck block of the hidden layer, residual connections are used to avoid information loss and instability, which is common in most modern neural network architectures. The residual connection also implicitly increases the sample numbers because the residual connection transfers the same input to the next layer to be processed by more filters. Therefore, we usually do not have to worry about its stability when using this method for hidden layers of a neural network with residual connections.

Essentially, our methods target towards processing the volume data that have dense, continuous 3D textures. To the best of our knowledge, such data mainly come from 3D computational imaging, such as 3D medical imaging and 3D seismic imaging. In contrast, the so-called volume data derived from point clouds are much sparser and mainly have surface textures, where rotational equivariance plays a major role, but due to the sparsity of data points, the higher radial orthogonality and affine equivariance of the filter basis may have only a limited impact on its processing.

As for the choice of application tasks, we focus on testing our method on the 3D image segmentation task, because image segmentation itself is an image generation task that is suitable for demonstrating transformation equivariance performance according to the definition of group equivariance. Similarly, we can also apply our methods to volumetric image restoration, computational imaging, or other volumetric image generation tasks, which will be tested in our future work. In contrast, image classification is mainly related to transformation invariance. However, as we have seen the successful application of WMCG-CNN in 2D image classification, our method should also be applicable to 3D image classification. We will explore this in future work.

In general, our work embodies the effort for a more interpretable deep learning. This is because, essentially, a conventional deep CNN can be considered as a WMCG-CNN using Kronecker delta bases. Therefore, we can interpret the 3D CNNs from the perspective of group equivariance. Our work increases the flexibility in designing deep 3D CNNs. The choice of filter bases becomes another "hyperparameter" that can be used to control the performance of the deep learning model.

\appendix

\section{Proof of Theorem \ref{theorem:decomp}}
\label{app1}
\setcounter{equation}{0}
\renewcommand{\thesection}{\Alph{section}}
\numberwithin{equation}{section}

\begin{proof}
First, we know that $M(a)$'s determinant $|M(a)|=2^{\alpha+\beta+\gamma}>0$. Thus, all the 3D matrices generated by \eqref{eq:matrixdecomp} belongs to $GL^+(3,\mathbb{R})$.

Second, considering any 3D matrix $\mathbf{Y}$ with a positive determinant, we need to find a suitable vector $a$ so that $M(a)=\mathbf{Y}$, such decomposition is equivalent to its inverse $M^{-1}(a)=\mathbf{Y}^{-1}$.

According to the properties of shear transform matrices, it is easier to know that by choosing suitable shear values (ie., $s_{01},s_{10},s_{02},s_{20},s_{12},s_{21}$), we can have 
\begin{equation}
\begin{aligned}
	&S_{20}(s_{20})S_{10}(s_{10})S_{21}(s_{21})S_{01}(s_{01})\cdot\\
	&S_{12}(s_{12})S_{02}(s_{02})\mathbf{Y}^{-1} = \mathbf{D}
\end{aligned}
\label{eq:scaling1}
\end{equation}
where 
\begin{equation}
\begin{aligned}
	\mathbf{D}=
	\begin{bmatrix}
		a & 0 & 0\\
		0 & b & 0\\
		0 & 0 & c
	\end{bmatrix},
\end{aligned}
\label{eq:scaling1}
\end{equation}
and $abc>0$, $a,b,c\in \mathbb{R}$.

Therefore, for decomposing $Y$ into the form of \eqref{eq:matrixdecomp}, we only need to further have 
\begin{equation}
A_3(-\gamma)A_2(-\beta)A_1(-\alpha)R_3(-\theta_3)R_1(-\theta_1)=\mathbf{D}.
\label{eq:scaling1}
\end{equation}

When $a>0$, $b>0$, and $c>0$, we have 
$\alpha=-\log_2(a)$, $\beta=-\log_2(b)$, $\gamma=-\log_2(c)$, $\theta_1=0$, and $\theta_3=0$;

When $a>0$, $b<0$, and $c<0$, we have 
$\alpha=-\log_2(a)$, $\beta=-\log_2(-b)$, $\gamma=-\log_2(-c)$, $\theta_1=0$, and $\theta_3=\pi$;

When $a<0$, $b<0$, and $c>0$, we have 
$\alpha=-\log_2(-a)$, $\beta=-\log_2(-b)$, $\gamma=-\log_2(c)$, $\theta_1=\pi$, and $\theta_3=0$;

When $a<0$, $b>0$, and $c<0$, we have 
$\alpha=-\log_2(-a)$, $\beta=-\log_2(b)$, $\gamma=-\log_2(-c)$, $\theta_1=\pi$, and $\theta_3=\pi$.

The proof is complete.

\end{proof}

\section{The computational complexity of different 3D image segmentation neural networks}
\label{app2}
\setcounter{table}{0}
\renewcommand{\thetable}{B.\arabic{table}}

Table \ref{tab-extend-param} shows the Training VRAM Footprint (MB), Inference Time (s) per image, Params (M), and MACs (G) of each neural network on the BTCV dataset. The Training VRAM Footprint (MB) is tested on an A100 (40GB) GPU. For stability, the Inference Time (s) per image is tested on a workstation with an AMD Ryzen 9 9950X CPU and an RTX4090D GPU. We find that WMCG-sFB slightly increases the VRAM footprint during the training phase compared to vanilla CNN, and maintains the same inference speed with a reduced number of learnable parameters and the same computational complexity during the inference phase.

\begin{table*}[ht]
\caption{The extensive comparison results with the state-of-the-art models on the computational complexity and inference time.}
\label{table}
\setlength{\tabcolsep}{3pt}
\centering
\centerline{\resizebox{12.6cm}{!}{ 
	\begin{threeparttable}[b]
		\begin{tabular}{cccccc}
			\toprule
			Models                                            & Base Model                                                       & Training VRAM Footprint (MB) & Inference Time (s) & Params (M) & MACs (G) \\ \midrule
			nnUNet\cite{isensee2021nnu}      &                                                                  & 7542                        & 26                 & 16.5       & 186.6    \\ 
			nnUNet-Res-Enc-L\cite{isensee2024nnu}                                 &                                                                  & 10382                       & 31                 & 141.4      & 883.1    \\ 
			UNETR\cite{hatamizadeh2022unetr} &                                                                  & 5952                        & 23                 & 93.0       & 149.1    \\ 
			UNETR++\cite{shaker2022unetr++}  &                                                                  & 5904                        & 25                 & 43.0       & 77.3     \\ 
			UXNET\cite{lee20223d}            &                                                                  & 16204                       & 79                & 53.0       & 1159.1   \\ 
			UMamba\cite{U-Mamba}                                            &                                                                  & 23074                       & 107                & 42.9       & 836.0    \\ 
			SegMamba\cite{Xin_SegMamba_MICCAI2024}                                          &                                                                  & 15354                       & 75                & 67.4       & 1218.3   \\ 
			VSmTrans\cite{liu2024vsmtrans}                                          &                                                                  & 29470                       & 94                & 50.2       & 681.6    \\ 
			STUNet-L\cite{huang2023stu}                                          &                                                                  & 26114                       & 141                & 440.1      & 4811.6   \\ \midrule
			CNN                                               & \multirow{4}{*}{MedNeXt-S\cite{roy2023mednext}} & 20258                       & 89                & 6.0        & 141.9    \\ 
			SE(3)\cite{weiler20183d}                                             &                                                                  & 20282                       & 90                & 5.6        & 141.9    \\ 
			PDO\cite{he2023neural}                                               &                                                                  & 21484                       & 68                 & 7.4        & 477.4    \\ 
			WMCG-sFB                                          &                                                                  & 20370                       & 89                & 5.7        & 141.9    \\ \midrule
			CNN                                               & \multirow{4}{*}{MedNeXt-L\cite{roy2023mednext}} & 38434                       & 150                & 63.0       & 473.7    \\ 
			SE(3)\cite{weiler20183d}                                             &                                                                  & 38856                       & 154                & 61.6       & 473.7    \\ 
			PDO\cite{he2023neural}                                               &                                                                  & 38912                       & 118                & 68.5       & 1215.4   \\ 
			WMCG-sFB                                          &                                                                  & 38864                       & 150                & 61.9       & 473.7    \\ \bottomrule
		\end{tabular}
		\end{threeparttable}}}
		\label{tab-extend-param}
\end{table*}

\bibliographystyle{unsrtnat}
\bibliography{references}

@inproceedings{azad2024beyond,
	title={Beyond self-attention: Deformable large kernel attention for medical image segmentation},
	author={Azad, Reza and Niggemeier, Leon and H{\"u}ttemann, Michael and Kazerouni, Amirhossein and Aghdam, Ehsan Khodapanah and Velichko, Yury and Bagci, Ulas and Merhof, Dorit},
	booktitle={Proceedings of the IEEE/CVF Winter Conference on Applications of Computer Vision},
	pages={1287--1297},
	year={2024}
}

@article{cohen2016steerable,
	title={Steerable cnns},
	author={Cohen, Taco S and Welling, Max},
	journal={arXiv preprint arXiv:1612.08498},
	year={2016}
}

@article{thomas2018tensor,
	title={Tensor field networks: Rotation-and translation-equivariant neural networks for 3d point clouds},
	author={Thomas, Nathaniel and Smidt, Tess and Kearnes, Steven and Yang, Lusann and Li, Li and Kohlhoff, Kai and Riley, Patrick},
	journal={arXiv preprint arXiv:1802.08219},
	year={2018}
}

@article{shen2024rotation,
	title={Rotation-Equivariant Quaternion Neural Networks for 3D Point Cloud Processing},
	author={Shen, Wen and Wei, Zhihua and Ren, Qihan and Zhang, Binbin and Huang, Shikun and Fan, Jiaqi and Zhang, Quanshi},
	journal={IEEE Transactions on Pattern Analysis and Machine Intelligence},
	year={2024},
	publisher={IEEE}
}

@article{weiler20183d,
	title={3d steerable cnns: Learning rotationally equivariant features in volumetric data},
	author={Weiler, Maurice and Geiger, Mario and Welling, Max and Boomsma, Wouter and Cohen, Taco S},
	journal={Advances in Neural Information Processing Systems},
	volume={31},
	year={2018}
}

@inproceedings{cesa2021program,
	title={A program to build E (N)-equivariant steerable CNNs},
	author={Cesa, Gabriele and Lang, Leon and Weiler, Maurice},
	booktitle={International Conference on Learning Representations},
	year={2021}
}

@inproceedings{zhu2023e2pn,
	title={E2PN: Efficient SE (3)-equivariant point network},
	author={Zhu, Minghan and Ghaffari, Maani and Clark, William A and Peng, Huei},
	booktitle={Proceedings of the IEEE/CVF Conference on Computer Vision and Pattern Recognition},
	pages={1223--1232},
	year={2023}
}

@inproceedings{zhdanov2023implicit,
	title={Implicit Convolutional Kernels for Steerable CNNs},
	author={Zhdanov, Maksim and Hoffmann, Nico and Cesa, Gabriele},
	booktitle={Thirty-seventh Conference on Neural Information Processing Systems},
	year={2023}
}

@inproceedings{lee2023scaling,
	title={Scaling up 3d kernels with bayesian frequency re-parameterization for medical image segmentation},
	author={Lee, Ho Hin and Liu, Quan and Bao, Shunxing and Yang, Qi and Yu, Xin and Cai, Leon Y and Li, Thomas Z and Huo, Yuankai and Koutsoukos, Xenofon and Landman, Bennett A},
	booktitle={International Conference on Medical Image Computing and Computer-Assisted Intervention},
	pages={632--641},
	year={2023},
	organization={Springer}
}

@inproceedings{roy2023mednext,
	title={Mednext: transformer-driven scaling of convnets for medical image segmentation},
	author={Roy, Saikat and Koehler, Gregor and Ulrich, Constantin and Baumgartner, Michael and Petersen, Jens and Isensee, Fabian and Jaeger, Paul F and Maier-Hein, Klaus H},
	booktitle={International Conference on Medical Image Computing and Computer-Assisted Intervention},
	pages={405--415},
	year={2023},
	organization={Springer}
}

@inproceedings{kondor2018generalization,
	title={On the generalization of equivariance and convolution in neural networks to the action of compact groups},
	author={Kondor, Risi and Trivedi, Shubhendu},
	booktitle={International Conference on Machine Learning},
	pages={2747--2755},
	year={2018},
	organization={PMLR}
}

@article{zhao2023adaptive,
	title={Adaptive aggregation of Monte Carlo augmented decomposed filters for efficient group-equivariant convolutional neural network},
	author={Zhao, Wenzhao and Wichtmann, Barbara D and Albert, Steffen and Maurer, Angelika and Z{\"o}llner, Frank G and Attenberger, Ulrike and Hesser, J{\"u}rgen},
	journal={arXiv preprint arXiv:2305.10110},
	year={2023}
}

@article{binney1991gaussian,
	title={Gaussian random fields in spherical coordinates},
	author={Binney, James and Quinn, Thomas},
	journal={Monthly Notices of the Royal Astronomical Society},
	volume={249},
	number={4},
	pages={678--683},
	year={1991},
	publisher={Oxford University Press Oxford, UK}
}

@inproceedings{xie2018orthogonality,
	title={Orthogonality-promoting distance metric learning: convex relaxation and theoretical analysis},
	author={Xie, Pengtao and Wu, Wei and Zhu, Yichen and Xing, Eric},
	booktitle={International Conference on Machine Learning},
	pages={5403--5412},
	year={2018},
	organization={PMLR}
}

@article{fisher1995wiener,
	title={Wiener reconstruction of density, velocity and potential fields from all-sky galaxy redshift surveys},
	author={Fisher, Karl B and Lahav, Ofer and Hoffman, Yehuda and Lynden-Bell, Donald and Zaroubi, Saleem},
	journal={Monthly Notices of the Royal Astronomical Society},
	volume={272},
	number={4},
	pages={885--908},
	year={1995},
	publisher={Oxford University Press Oxford, UK}
}

@inproceedings{roth2015deeporgan,
	title={Deeporgan: Multi-level deep convolutional networks for automated pancreas segmentation},
	author={Roth, Holger R and Lu, Le and Farag, Amal and Shin, Hoo-Chang and Liu, Jiamin and Turkbey, Evrim B and Summers, Ronald M},
	booktitle={Medical Image Computing and Computer-Assisted Intervention--MICCAI 2015: 18th International Conference, Munich, Germany, October 5-9, 2015, Proceedings, Part I 18},
	pages={556--564},
	year={2015},
	organization={Springer}
}

@inproceedings{hatamizadeh2022unetr,
	title={Unetr: Transformers for 3d medical image segmentation},
	author={Hatamizadeh, Ali and Tang, Yucheng and Nath, Vishwesh and Yang, Dong and Myronenko, Andriy and Landman, Bennett and Roth, Holger R and Xu, Daguang},
	booktitle={Proceedings of the IEEE/CVF winter conference on applications of computer vision},
	pages={574--584},
	year={2022}
}

@article{shaker2022unetr++,
	title={UNETR++: delving into efficient and accurate 3D medical image segmentation},
	author={Shaker, Abdelrahman and Maaz, Muhammad and Rasheed, Hanoona and Khan, Salman and Yang, Ming-Hsuan and Khan, Fahad Shahbaz},
	journal={arXiv preprint arXiv:2212.04497},
	year={2022}
}

@article{shaker2024unetr++,
	title={UNETR++: delving into efficient and accurate 3D medical image segmentation},
	author={Shaker, Abdelrahman M and Maaz, Muhammad and Rasheed, Hanoona and Khan, Salman and Yang, Ming-Hsuan and Khan, Fahad Shahbaz},
	journal={IEEE Transactions on Medical Imaging},
	year={2024},
	publisher={IEEE}
}

@article{lee20223d,
	title={3d ux-net: A large kernel volumetric convnet modernizing hierarchical transformer for medical image segmentation},
	author={Lee, Ho Hin and Bao, Shunxing and Huo, Yuankai and Landman, Bennett A},
	journal={arXiv preprint arXiv:2209.15076},
	year={2022}
}

@article{isensee2021nnu,
	title={nnU-Net: a self-configuring method for deep learning-based biomedical image segmentation},
	author={Isensee, Fabian and Jaeger, Paul F and Kohl, Simon AA and Petersen, Jens and Maier-Hein, Klaus H},
	journal={Nature methods},
	volume={18},
	number={2},
	pages={203--211},
	year={2021},
	publisher={Nature Publishing Group}
}

@article{fukushima1980neocognitron,
	title={Neocognitron: A self-organizing neural network model for a mechanism of pattern recognition unaffected by shift in position},
	author={Fukushima, Kunihiko},
	journal={Biological cybernetics},
	volume={36},
	number={4},
	pages={193--202},
	year={1980},
	publisher={Springer}
}

@article{lecun1989backpropagation,
	title={Backpropagation applied to handwritten zip code recognition},
	author={LeCun, Yann and Boser, Bernhard and Denker, John S and Henderson, Donnie and Howard, Richard E and Hubbard, Wayne and Jackel, Lawrence D},
	journal={Neural computation},
	volume={1},
	number={4},
	pages={541--551},
	year={1989},
	publisher={MIT Press}
}

@inproceedings{cohen2016group,
	title={Group equivariant convolutional networks},
	author={Cohen, Taco and Welling, Max},
	booktitle={International conference on machine learning},
	pages={2990--2999},
	year={2016},
	organization={PMLR}
}

@inproceedings{sangalli2021scale,
	title={Scale equivariant neural networks with morphological scale-spaces},
	author={Sangalli, Mateus and Blusseau, Samy and Velasco-Forero, Santiago and Angulo, Jesus},
	booktitle={Discrete Geometry and Mathematical Morphology: First International Joint Conference, DGMM 2021, Uppsala, Sweden, May 24--27, 2021, Proceedings},
	pages={483--495},
	year={2021},
	organization={Springer}
}

@article{gao2021deformation,
	title={Deformation robust roto-scale-translation equivariant cnns},
	author={Gao, Liyao and Lin, Guang and Zhu, Wei},
	journal={arXiv preprint arXiv:2111.10978},
	year={2021}
}

@article{sosnovik2019scale,
	title={Scale-equivariant steerable networks},
	author={Sosnovik, Ivan and Szmaja, Micha{\l} and Smeulders, Arnold},
	journal={arXiv preprint arXiv:1910.11093},
	year={2019}
}

@article{sosnovik2021disco,
	title={DISCO: accurate Discrete Scale Convolutions},
	author={Sosnovik, Ivan and Moskalev, Artem and Smeulders, Arnold},
	journal={arXiv preprint arXiv:2106.02733},
	year={2021}
}

@article{zhu2022scaling,
	title={Scaling-Translation-Equivariant Networks with Decomposed Convolutional Filters},
	author={Zhu, Wei and Qiu, Qiang and Calderbank, Robert and Sapiro, Guillermo and Cheng, Xiuyuan},
	journal={Journal of machine learning research},
	volume={23},
	number={68},
	pages={1--45},
	year={2022}
}

@inproceedings{landman2015miccai,
	title={Miccai multi-atlas labeling beyond the cranial vault--workshop and challenge},
	author={Landman, Bennett and Xu, Zhoubing and Igelsias, J and Styner, Martin and Langerak, T and Klein, Arno},
	booktitle={Proc. MICCAI Multi-Atlas Labeling Beyond Cranial Vault—Workshop Challenge},
	volume={5},
	pages={12},
	year={2015}
}

@inproceedings{liu2022convnet,
	title={A convnet for the 2020s},
	author={Liu, Zhuang and Mao, Hanzi and Wu, Chao-Yuan and Feichtenhofer, Christoph and Darrell, Trevor and Xie, Saining},
	booktitle={Proceedings of the IEEE/CVF conference on computer vision and pattern recognition},
	pages={11976--11986},
	year={2022}
}

@article{soler20103d,
	title={3D image reconstruction for comparison of algorithm database},
	author={Soler, Luc and Hostettler, Alexandre and Agnus, Vincent and Charnoz, Arnaud and Fasquel, Jean-Baptiste and Moreau, Johan and Osswald, Anne-Blandine and Bouhadjar, Mourad and Marescaux, Jacques},
	journal={URL: https://www. ircad. fr/research/data-sets/liver-segmentation-3d-ircadb-01},
	year={2010}
}

@article{baid2021rsna,
	title={The rsna-asnr-miccai brats 2021 benchmark on brain tumor segmentation and radiogenomic classification},
	author={Baid, Ujjwal and Ghodasara, Satyam and Mohan, Suyash and Bilello, Michel and Calabrese, Evan and Colak, Errol and Farahani, Keyvan and Kalpathy-Cramer, Jayashree and Kitamura, Felipe C and Pati, Sarthak and others},
	journal={arXiv preprint arXiv:2107.02314},
	year={2021}
}

@article{U-Mamba,
	title={U-Mamba: Enhancing Long-range Dependency for Biomedical Image Segmentation},
	author={Ma, Jun and Li, Feifei and Wang, Bo},
	journal={arXiv preprint arXiv:2401.04722},
	year={2024}
}

@InProceedings{Xin_SegMamba_MICCAI2024,
	author = { Xing, Zhaohu and Ye, Tian and Yang, Yijun and Liu, Guang and Zhu, Lei},
	title = { { SegMamba: Long-range Sequential Modeling Mamba For 3D Medical Image Segmentation } },
	booktitle = {proceedings of Medical Image Computing and Computer Assisted Intervention -- MICCAI 2024},
	year = {2024},
	publisher = {Springer Nature Switzerland},
	volume = {LNCS 15008},
	month = {October},
	page = {578 -- 588}
}

@article{liu2024vsmtrans,
	title={VSmTrans: A hybrid paradigm integrating self-attention and convolution for 3D medical image segmentation},
	author={Liu, Tiange and Bai, Qingze and Torigian, Drew A and Tong, Yubing and Udupa, Jayaram K},
	journal={Medical Image Analysis},
	volume={98},
	pages={103295},
	year={2024},
	publisher={Elsevier}
}

@inproceedings{
	he2023neural,
	title={Neural e{PDO}s: Spatially Adaptive Equivariant Partial Differential Operator Based  Networks},
	author={Lingshen He and Yuxuan Chen and Zhengyang Shen and Yibo Yang and Zhouchen Lin},
	booktitle={The Eleventh International Conference on Learning Representations },
	year={2023},
	url={https://openreview.net/forum?id=D1Iqfm7WTkk}
}

@article{jiao2019survey,
	title={A survey on the new generation of deep learning in image processing},
	author={Jiao, Licheng and Zhao, Jin},
	journal={Ieee Access},
	volume={7},
	pages={172231--172263},
	year={2019},
	publisher={IEEE}
}

@inproceedings{ronneberger2015u,
	title={U-net: Convolutional networks for biomedical image segmentation},
	author={Ronneberger, Olaf and Fischer, Philipp and Brox, Thomas},
	booktitle={Medical image computing and computer-assisted intervention--MICCAI 2015: 18th international conference, Munich, Germany, October 5-9, 2015, proceedings, part III 18},
	pages={234--241},
	year={2015},
	organization={Springer}
}

@article{chen2024hidiff,
	title={HiDiff: Hybrid Diffusion Framework for Medical Image Segmentation},
	author={Chen, Tao and Wang, Chenhui and Chen, Zhihao and Lei, Yiming and Shan, Hongming},
	journal={IEEE Transactions on Medical Imaging},
	year={2024},
	publisher={IEEE}
}

@article{shen2024efficient,
	title={Efficient learning of Scale-Adaptive Nearly Affine Invariant Networks},
	author={Shen, Zhengyang and Qiu, Yeqing and Liu, Jialun and He, Lingshen and Lin, Zhouchen},
	journal={Neural Networks},
	volume={174},
	pages={106229},
	year={2024},
	publisher={Elsevier}
}

@article{mo2024ric,
	title={RIC-CNN: rotation-invariant coordinate convolutional neural network},
	author={Mo, Hanlin and Zhao, Guoying},
	journal={Pattern Recognition},
	volume={146},
	pages={109994},
	year={2024},
	publisher={Elsevier}
}

@article{zhong2022pesa,
	title={PESA-Net: Permutation-equivariant split attention network for correspondence learning},
	author={Zhong, Zhen and Xiao, Guobao and Wang, Shiping and Wei, Leyi and Zhang, Xiaoqin},
	journal={Information Fusion},
	volume={77},
	pages={81--89},
	year={2022},
	publisher={Elsevier}
}

@inproceedings{cheng2019rotdcf,
	title={ROTDCF: DECOMPOSITION OF CONVOLUTIONAL FILTERS FOR ROTATION-EQUIVARIANT DEEP NETWORKS},
	author={Cheng, X and Qiu, Q and Calderbank, R and Sapiro, G},
	booktitle={International Conference Learning Representation},
	year={2019}
}

@article{qiu2018dcfnet,
	title={{DCFNet}: Deep Neural Network with Decomposed Convolutional Filters},
	author={Qiu, Qiang and Cheng, Xiuyuan and Calderbank, Robert and Sapiro, Guillermo},
	journal={International Conference on Machine Learning},
	year={2018}
}

@article{alaudah2019machine,
	title={A machine-learning benchmark for facies classification},
	author={Alaudah, Yazeed and Micha{\l}owicz, Patrycja and Alfarraj, Motaz and AlRegib, Ghassan},
	journal={Interpretation},
	volume={7},
	number={3},
	pages={SE175--SE187},
	year={2019},
	publisher={Society of Exploration Geophysicists and American Association of Petroleum}
}

@ARTICLE{9686703,
	author={Chai, Xintao and Nie, Wenhui and Lin, Kai and Tang, Genyang and Yang, Taihui and Yu, Junyong and Cao, Wenjun},
	journal={IEEE Transactions on Geoscience and Remote Sensing}, 
	title={An Open-Source Package for Deep-Learning-Based Seismic Facies Classification: Benchmarking Experiments on the SEG 2020 Open Data}, 
	year={2022},
	volume={60},
	number={},
	pages={1-19},
	doi={10.1109/TGRS.2022.3144666}
}

@article{ORE2025105823,
	title = {SeisSegDiff: A label-efficient few-shot texture segmentation diffusion model for seismic facies classification},
	journal = {Computers \& Geosciences},
	volume = {196},
	pages = {105823},
	year = {2025},
	issn = {0098-3004},
	author = {Tobi Ore and Dengliang Gao}
}

@book{dummit2004abstract,
	title={Abstract algebra},
	author={Dummit, David Steven and Foote, Richard M and others},
	volume={3},
	year={2004},
	publisher={Wiley Hoboken}
}

@book{gurarie2007symmetries,
	title={Symmetries and Laplacians: introduction to harmonic analysis, group representations and applications},
	author={Gurarie, David},
	year={2007},
	publisher={Courier Corporation}
}

@inproceedings{isensee2024nnu,
	title={nnu-net revisited: A call for rigorous validation in 3d medical image segmentation},
	author={Isensee, Fabian and Wald, Tassilo and Ulrich, Constantin and Baumgartner, Michael and Roy, Saikat and Maier-Hein, Klaus and Jaeger, Paul F},
	booktitle={International Conference on Medical Image Computing and Computer-Assisted Intervention},
	pages={488--498},
	year={2024},
	organization={Springer}
}

@article{huang2023stu,
	title={Stu-net: Scalable and transferable medical image segmentation models empowered by large-scale supervised pre-training},
	author={Huang, Ziyan and Wang, Haoyu and Deng, Zhongying and Ye, Jin and Su, Yanzhou and Sun, Hui and He, Junjun and Gu, Yun and Gu, Lixu and Zhang, Shaoting and others},
	journal={arXiv preprint arXiv:2304.06716},
	year={2023}
}

@inproceedings{10.5555/3600270.3602415,
	author = {You, Chenyu and Zhao, Ruihan and Liu, Fenglin and Dong, Siyuan and Chinchali, Sandeep and Topcu, Ufuk and Staib, Lawrence and Duncan, James S.},
	title = {Class-aware adversarial transformers for medical image segmentation},
	year = {2022},
	isbn = {9781713871088},
	publisher = {Curran Associates Inc.},
	address = {Red Hook, NY, USA},
	booktitle = {Proceedings of the 36th International Conference on Neural Information Processing Systems},
	articleno = {2145},
	numpages = {15},
	location = {New Orleans, LA, USA},
	series = {NIPS '22}
}

@inproceedings{you2023bootstrapping,
	title={Bootstrapping semi-supervised medical image segmentation with anatomical-aware contrastive distillation},
	author={You, Chenyu and Dai, Weicheng and Min, Yifei and Staib, Lawrence and Duncan, James S},
	booktitle={International conference on information processing in medical imaging},
	pages={641--653},
	year={2023},
	organization={Springer}
}

@article{you2024mine,
	title={Mine your own anatomy: Revisiting medical image segmentation with extremely limited labels},
	author={You, Chenyu and Dai, Weicheng and Liu, Fenglin and Min, Yifei and Dvornek, Nicha C and Li, Xiaoxiao and Clifton, David A and Staib, Lawrence and Duncan, James S},
	journal={IEEE Transactions on Pattern Analysis and Machine Intelligence},
	year={2024},
	publisher={IEEE}
}

@article{you2023rethinking,
	title={Rethinking semi-supervised medical image segmentation: A variance-reduction perspective},
	author={You, Chenyu and Dai, Weicheng and Min, Yifei and Liu, Fenglin and Clifton, David and Zhou, S Kevin and Staib, Lawrence and Duncan, James},
	journal={Advances in neural information processing systems},
	volume={36},
	pages={9984--10021},
	year={2023}
}

@inproceedings{you2023action++,
	title={Action++: Improving semi-supervised medical image segmentation with adaptive anatomical contrast},
	author={You, Chenyu and Dai, Weicheng and Min, Yifei and Staib, Lawrence and Sekhon, Jas and Duncan, James S},
	booktitle={International Conference on Medical Image Computing and Computer-Assisted Intervention},
	pages={194--205},
	year={2023},
	organization={Springer}
}

@inproceedings{you2023implicit,
	title={Implicit anatomical rendering for medical image segmentation with stochastic experts},
	author={You, Chenyu and Dai, Weicheng and Min, Yifei and Staib, Lawrence and Duncan, James S},
	booktitle={International Conference on Medical Image Computing and Computer-Assisted Intervention},
	pages={561--571},
	year={2023},
	organization={Springer}
}

\end{document}